# AI Enabling Technologies: A Survey

Vijay Gadepally, Justin Goodwin, Jeremy Kepner, Albert Reuther, Hayley Reynolds, Siddharth Samsi, Jonathan Su, David Martinez

MIT Lincoln Laboratory
244 Wood Street
Lexington, MA, 02421

## ABSTRACT

Artificial intelligence (AI) has the opportunity to revolutionize the way the United States Department of Defense (DoD) and Intelligence Community (IC) address the challenges of evolving threats, data deluge, and rapid courses of action. Developing an end-to-end AI system involves parallel development of different pieces that must work together in order to provide capabilities that can be used by decision makers, warfighters, and analysts. These pieces include data collection, data conditioning, algorithms, computing, robust AI, and human–machine teaming. Although much of the popular press today surrounds advances in algorithms and computing, most modern AI systems leverage advances across numerous different fields. Further, while certain components may not be as visible to end-users as others, our experience has shown that each of these interrelated components play a major role in the success or failure of an AI system. This article is meant to highlight many of these technologies that are involved in an end-to-end AI system. The goal of this article is to provide readers with an overview of terminology, technical details, and recent highlights from academia, industry, and government. Where possible, we indicate relevant resources that can be used for further reading and understanding.

This material is based upon work supported by the United States Air Force under Air Force Contract No. FA8702-15-D-0001. Any opinions, findings, conclusions or recommendations expressed in this material are those of the author(s) and do not necessarily reflect the views of the United States Air Force.

# 1 INTRODUCTION

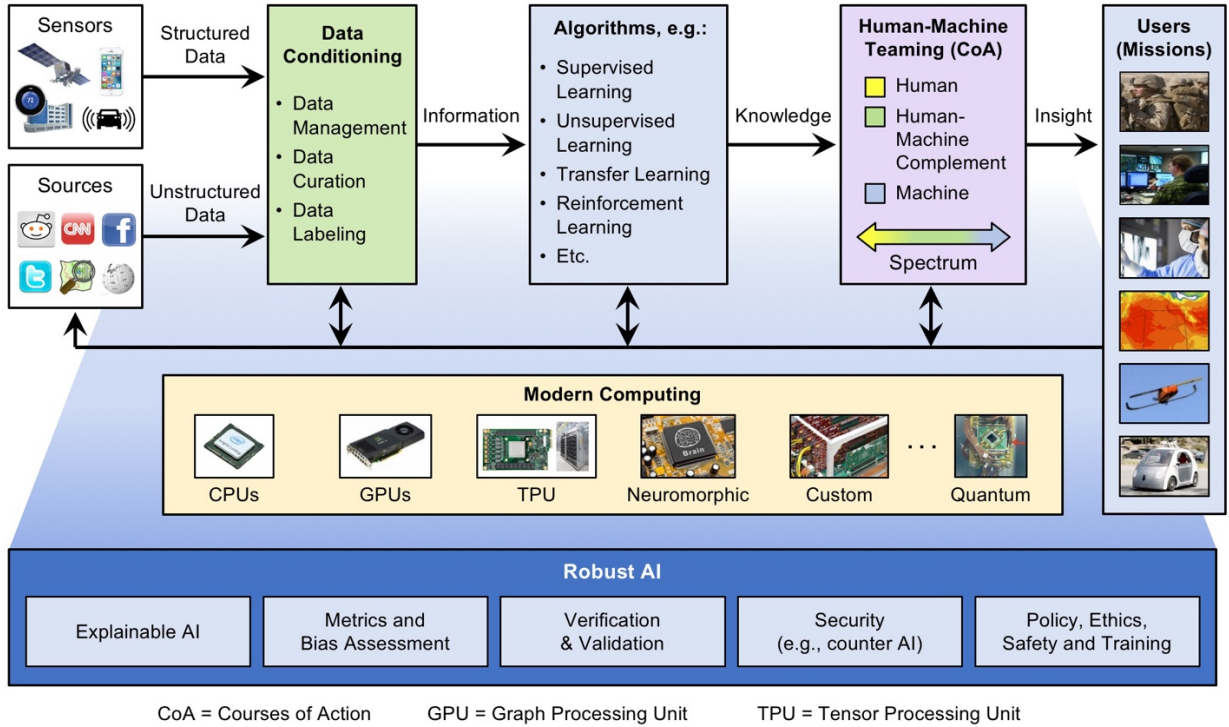

*Figure 1.1. Canonical AI architecture consists of sensors, data conditioning, algorithms, modern computing, robust AI, human–machine teaming, and users (missions). Each step is critical in developing end-to-end AI applications and systems.*

AI has the opportunity to revolutionize the way the DoD and ICs address the challenges of evolving threats, data deluge, and rapid courses of action. AI solutions involve a number of different pieces that must work together in order to provide capabilities that can be used by decision makers, warfighters, and analysts. Consider the canonical architecture of an AI system in Figure 1.1. This figure outlines many of the important components needed when developing an end-to-end AI solution. While much of the popular press surrounds advances in algorithms and computing, most modern AI systems leverage advances across numerous different fields. Further, while certain components may not be as visible to end users as others, our experience has shown that each of these interrelated components play a major role in the success or failure of an AI system.

On the left side of Figure 1.1, we have data coming in from a variety of structured and unstructured sources. Often, these *structured and unstructured data* sources together provide different views of the same entities and/or phenomenology. Often data from sensors are categorized as structured data because the raw digital data are accompanied by metadata. In contrast, "unstructured data" is how we refer to data in which there is no predefined structure or source of metadata contained.

These raw data are often fed into a *data conditioning* step in which they are fused, aggregated, structured, accumulated, and converted to information. The main objective for this subcomponent is to transform data into information. An example of information is a new sensor image (after data labeling) that we need to use to classify if the object of interest is present in that image or not (like a vehicle of interest). Typical functions performed under this subcomponent are: standardization of data formats complying with a data ontology, data labeling, highlights of missing or incomplete data, errors/biases in the data, etc.

The information generated by the data conditioning step feeds into a host of supervised and unsupervised algorithms such as neural networks. These algorithms are used to extract patterns, predict new events, fill in missing data, or look for similarities across datasets. These algorithms essentially convert the input information to actionable knowledge. In our definition, we use the term "knowledge" to describe information that has been converted into a higher-level representation that is ready for human consumption.

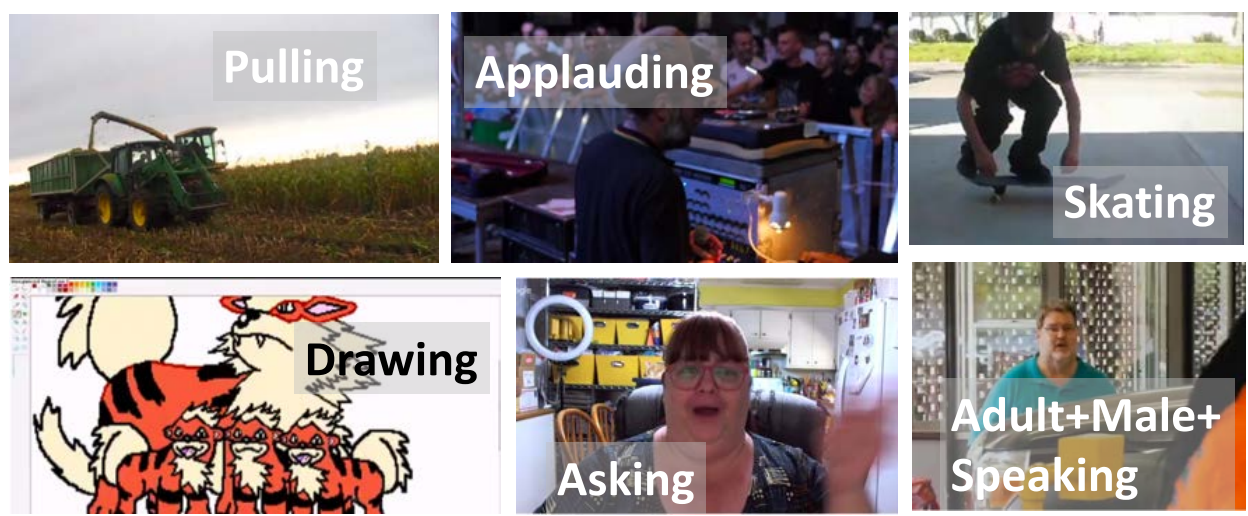

*Figure 1.2. Example categories and video screen shots from the Moments in Time Dataset.*

With the knowledge extracted in the algorithms phase, it is important to include the human being in the decision-making process. This is done in the *human–machine teaming* phase. Although there are a few applications that may be amenable to autonomous decision making (e.g., email spam filtering), recent AI advances of relevance to the DoD have largely been in fields where a human is either in- or on- the-loop. The phase of human–machine teaming is critical in connecting the data and algorithms to the end user and in providing the mission users with useful and relevant insight. Human–machine teaming is the phase in which knowledge can be turned into actionable intelligence or insight by effectively utilizing human and machine resources as appropriate.

Underpinning all of these phases is the bedrock of *modern computing* systems made up of a number of heterogenous computing elements. For example, sensor processing may occur on low power embedded computers whereas algorithms may be computed in very large data centers. With the end of Moore's law [1], we've seen a Cambrian explosion of computing technologies and architectures. Understanding the relative benefits of these technologies is of particular importance to applying AI to domains under significant constraints such as size, weight, and power.

Another foundational technology underpinning AI development is *robust or trusted AI*. In this area, researchers are looking at ways to explain AI outcomes (for example, why a system is recommending a particular course of action); metrics to measure the effectiveness of an AI algorithm (going beyond the traditional accuracy and precision metrics for complex applications or decisions); verification and validation (ensuring that results are provably correct under adversarial conditions); security (dealing with malicious or counter-AI technology); and policy decisions that govern the safe, responsible, and ethical use of AI technology. Although traditional academic and commercial players are looking at these issues, some non-profit initiatives such as OpenAI or the Allen Institute are taking a leading role in this area.

In the following sections, we highlight some of the salient technical concepts, research challenges, and opportunities for each of these core components of an AI system. In order to elucidate these components, we also use a running example based on research applying high-performance computing (HPC) to video classification. We would also like to note that each of the components of the AI architecture are vast academic areas with rich histories and numerous well published results. In order to provide readers with an overall view of all the components within this section, we concentrate on high-level concepts and also include vignettes of select research highlights or application examples.

## 1.1 VIDEO CLASSIFICATION EXAMPLE OVERVIEW

Over the course of this section, in order to provide concrete examples of components of the AI architecture being discussed, we use a running example based on our research of using high performance computing for video classification purposes. Specifically, we concentrate on the recently developed Moments in Time Dataset [2] developed at the Massachusetts Institute of Technology (MIT) Computer Science and Artificial Intelligence Laboratory (CSAIL). This dataset consists of one million videos given a label corresponding to an action being performed in the video. Each video is approximately three seconds in length and is labeled according to what a human observer believes is happening in the video. For example, a video of a dog barking is classified as "barking" and a video of people clapping would be labeled as "clapping." Figure 1.2 shows a few screenshots of videos from the dataset and associated labels. Of course, there are many areas where a particular label may not be as precise. For example, videos with the action label "cramming" could imply a person studying before an exam or someone putting something into a box. As of now, each video in the Moments in Time Dataset is labeled with one of approximately 380 possible labels. Some of the video clips also contain audio, but it is not necessarily present for all videos.

The Moments in Time Dataset is an example of a well-curated dataset that can be used to conduct research on video classification. To this effect, the creators of the dataset held a competition in 2018 to encourage dataset usage and share results that highlight the state of the field. Information about this competition can be found at: https://moments.csail.mit.edu/challenge2018/

As a metric to present the quality of a particular algorithm, the competition called for presentation of a top-*k* accuracy score. This metric is defined as follows: An algorithm will label each of the videos with one of *k* labels. The top-*k* accuracy says that a video was correctly identified if one of its top *k* labels is the correct label. For example, a video may be classified (in decreasing probability) as: *(barking, yelling, running, …)*. If the correct label (as judged by a human observer) is "yelling", the top-five accuracy for this would be 1. The top-one accuracy would be 0. As of June 2018, competition winners had top-one accuracies of approximately 0.3 and top-five accuracies of approximately 0.6 [3–5].

## 2 DATA CONDITIONING

Many AI application developers typically begin with a dataset of interest and a vision of the end analytic or insight they wish to gain from the data at hand. Although these are two very important components of the AI pipeline, one often spends the first few weeks (sometimes months) in the phase we refer to as data conditioning. This step typically includes tasks such as figuring out how to store data, dealing with inconsistencies in the dataset, and determining which algorithm (or set of algorithms) will be best suited for the application. Larger, faster, and messier datasets such as those from Internet of Things sensors, medical devices or autonomous vehicles only amplify these issues. These challenges, often referred to as the three Vs (volume, velocity, variety) of Big Data, require low-level tools for data management and data cleaning/pre-processing. In most applications, data can come from structured and/or unstructured sources and often includes inconsistencies, formatting differences, and a lack of ground-truth labels.

By some accounts, data conditioning can account for nearly 80% of the time consumed in developing a data science or AI application [6]. Within the realm of data conditioning, specific tasks include data discovery, data linkage, outlier detection, data management, and data labeling.

At a high level, the concept of data conditioning is the effort required to go from raw sensor data to information that can be used in further processing steps. Sometimes this phase is also referred to as data wrangling. Typically, each of these data conditioning tasks can be cumbersome, require significant domain knowledge, and represent a significant hurdle in developing an AI application. Many of the recent algorithmic advances have, in fact, occurred in areas where "conditioned" data can be found. For example, advances in image classification were largely driven by the availability of the ImageNet dataset [7], advances in handwriting recognition by the Modified National Institute of Standards and Technology (MNIST) dataset [8], and advances in video recognition by the Moments in Time dataset [2]. Other popular datasets such as CIFAR-10 [9], ATARI games [10], and Internet traces [11] have also played roles in advancing certain classes of algorithms and genres of applications.

There are a number of research efforts and organizations aiming to reduce the data conditioning barrier to entry. In this section, we will focus on three particular aspects of data conditioning: data management, data curation, and data labeling (for supervised learning). The input to this phase is typically raw data from heterogenous sources. This step aims to convert these data by aggregating them in a single place, designing a schema that relates all the components and often performs rudimentary anomaly detection/outlier detection. In the following subsections, we describe a few approaches to these tasks.

### 2.1 DATA MANAGEMENT

AI and machine-learning systems are highly dependent on access to consistent and formatted data. However, it is rare that a collection of sensors such as those used in the AI pipeline directly provide this information in a consistent manner. For example, in the video classification example, certain cameras may be turned off, may have different metadata, have different compression techniques, have different frame rates, have different color normalization schemes, etc. Further, fusing different pieces of data coming from disparate sources can be a major challenge—like fusing the information contained in audio streams with video streams in the video classification example. One of the first challenges is providing a uniform platform in which data can be fused and managed.

Traditionally, database systems are seen as the natural data management approach. A database is a collection of data and supporting data structure. Traditionally, databases are exposed to users via a database management system. Users interact with these database management systems to define new data structures, schemas (data organization), to update data, and retrieve data. Beyond databases, developers may store data as files leveraging parallel file systems such as Lustre [12]. For the remainder of this section, however, we will focus on database systems such as those shown in Figure 2.1.

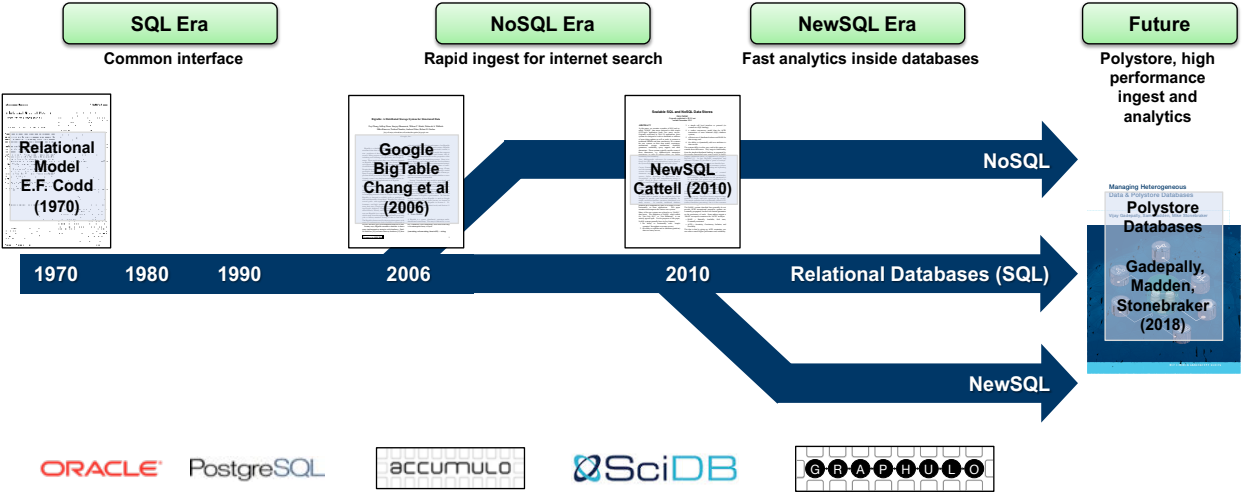

*Figure 2.1. Evolution of database management systems.*

Traditional database management systems such as Oracle [13] and PostGRES [14], sometimes referred to as relational databases, while compliant with ACID [15] guarantees, are unable to scale horizontally for certain applications [16]. To address these challenges, large internet companies such as Google and Facebook developed horizontally scalable database technologies such as BigTable [17] and Cassandra [18]. These NoSQL [19] (not-only structured query language [SQL]) technologies enabled rapid ingest and high performance even on relatively modest computing equipment. BigTable inspired databases such as Apache Accumulo [20] extended the NoSQL model for application specific requirements such as cell-level security. NoSQL databases do not provide the same level of guarantees on the data as relational databases [16]; however, they have been very popular due to their scalability, flexible data model, and tolerance to hardware failure. In the recent few years, spurred by inexpensive high performance hardware and custom hardware solutions, we have seen the evolution of a new era in database technologies, sometimes called NewSQL databases [21]. These data management systems largely support the scalability of NoSQL databases while preserving the data guarantees of SQL-era database systems. Largely, this is done by simplifying data models, such as in SciDB, or leveraging in-memory solutions such as in MemSQL and Spark. Looking towards the future, we see the development of new data management technologies that leverage the relative advantages of technologies developed within the various eras of database management technologies. A very high-level view of this evolution is presented in Figure 2.1. Looking towards the future, it is clear that no single type of database management systems is likely to support the kinds of data being collected from heterogenous sources of structured and unstructured data. Understanding how these different systems fundamentally interact with each other has a number of practical and theoretical [22], [23] implications. In order to address this challenge, one example of an active area of research in data management is in multi-database systems [24] such as Polystore databases and a specific example is the BigDAWG system described below.

### 2.1.1 Data Management Research Example—BigDAWG

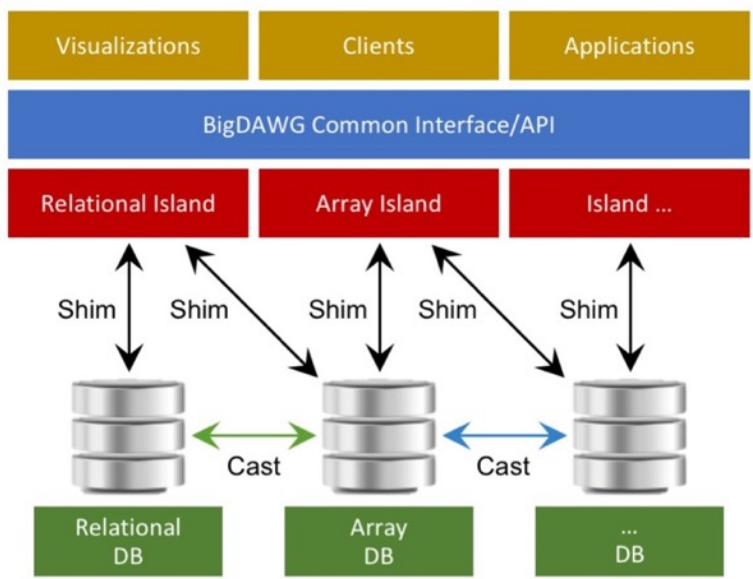

*Figure 2.2. BigDAWG architecture.*

BigDAWG [16, 25, 26], short for the Big Data Working Group, is an implementation of a polystore database system designed to simplify database management for complex applications. For example, modern decision support systems are required to integrate and synthesize a rapidly expanding collection of real-time data feeds: sensor data, analyst reports, social media, chat, documents, manifests, logistical data, and system logs (to name just a few). The traditional technique for solving a complex data fusion problem is to pick a single general-purpose database engine and move everything into this system. However, custom database engines for sensors, graphs, documents, and transactions (just to name a few) provide 100× better performance than general-purpose databases. The performance benefits of custom databases have resulted in the proliferation of data-specific databases, with most modern decision support systems containing five or more distinct customized storage systems. Additionally, for organizational or policy reasons, data may be required to stay in disparate database engines. For an application developer, this situation translates to developing his or her own interfaces and connectors for every different system. In general, for $N$ different systems, a user will have to create nearly $N^2$ different connectors. BigDAWG allows users to access data stored across multiple databases via a uniform common interface. Thus, for a complex application in which there is scientific data, text data, and metadata, a user can store each of these components in the storage technology best suited to each data type, but also develop analytics and applications that make use of all of these data without having to write custom connectors to each of these storage technologies. The end-to-end architecture of the BigDAWG polystore system is described in Figure 2.2. This architecture describes how applications, visualizations, and clients at the top access information stored in a variety of database engines at the bottom. At the bottom, we have a collection of disparate storage engines (we make no assumption about the data model, programming model, etc., of each of these engines). These storage engines are organized into a number of *islands*. An island is composed of a data model, a set of operations, and a set of candidate storage engines. An island provides location independence among its associated storage engines. A *shim* connects an island to one or more storage engines. The shim is basically a translator that maps queries expressed in terms of the operations defined by an island into the native query language of a particular storage engine. A key goal of a polystore system is for the processing to occur on the storage engine best suited to the features of the data. We expect in typical workloads that queries will produce results best suited to particular storage engines. Hence, BigDAWG needs a capability to move data directly between storage engines. We do this with software components we call *casts*.

### 2.1.2 Database and Storage Engines

A key design feature of BigDAWG is the support of multiple database and storage engines. With the rapid increase in heterogeneous data and the proliferation of highly specialized, tuned, and hardware-accelerated database engines, it is important that BigDAWG support as many data models as possible. Further, many organizations already rely on legacy systems as a part of their overall solution. We believe that analytics of the future will depend on many distinct data sources that can be efficiently stored and processed only in disparate systems. BigDAWG is designed to address this need by leveraging many vertically integrated data management systems. The current implementation of BigDAWG supports a number of popular database engines: PostGRES (SQL), MySQL (SQL), Vertica (SQL), Accumulo (NoSQL), SciDB (NewSQL), and S-Store (NewSQL). The modular design allows users to continue to integrate new engines as needed.

### 2.1.3 BigDAWG Islands

The next layer of the BigDAWG stack is its islands. Islands allow users to trade off between semantic completeness (using the full power of an underlying database engine) and location transparency (the ability to access data without knowledge of the underlying engine). Each island has a data model, a query language or set of operators, and one or more database engines for executing them. In the BigDAWG prototype, users determine the *scope* of their query by specifying an island within which the query will be executed. Islands are a user-facing abstraction, and they are designed to reduce the challenges associated with incorporating a new database engine. The current implementation of BigDAWG supports islands with relational, array, text, and streaming models. Our modular design supports the creation of new islands that encapsulate different programming and data models.

### 2.1.4 BigDAWG Middleware and API

The BigDAWG "secret sauce" lies in the middleware that is responsible for developing cross-engine query plans, monitoring previous queries and performance, migrating data across database engines as needed, and physically executing the requested query or analytic. The BigDAWG interface provides an API to execute polystore queries. The API layer consists of server- and client-facing components. The server components incorporate islands that connect to database engines via lightweight connectors referred to as shims. Shims essentially act as an adapter to go from the language of an island to the native language of an underlying database engine. In order to identify how a user is interacting with an island, a user specifies a scope in the query. A scope of a query allows an island to correctly interpret the syntax of the query and allows the island to select the correct shim that is needed to execute a part of the query. Thus, a cross-island query may involve multiple scope operations. Details of the BigDAWG middleware can be found in [27]–[30].

## 2.2 DATA CURATION

Data curation, within the context of our AI architecture, is used to refer to the process of maintaining data from creation (sensor collection) to usage in the machine-learning algorithm. In contrast to data management in the previous section, data curation typically involves some form of data normalization, data cleaning, data enrichment, and/or data discovery. Figure 2.3 describes where such curation activities may sit in relation to the data management and polystore technologies described in the previous section.

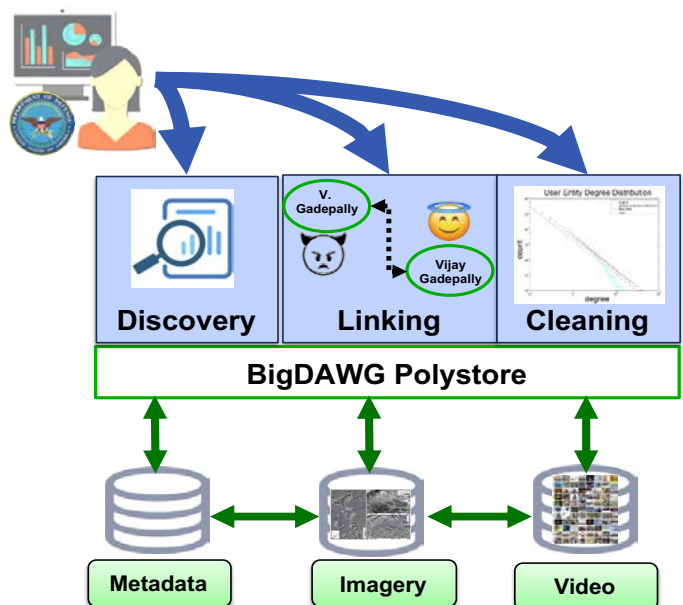

*Figure 2.3. Notional overview of data curation activities on top of data management.*

Although this is often a largely manual process with a human expert looking at data and determining whether there are errors, there are machine-learning techniques that can significantly reduce the amount of manual work needed. To do this, one may employ some form of unsupervised learning to look for obvious anomalies and outliers in the data. For example, in a medical dataset, one may incorrectly encode a field that is supposed to use inches as feet (thus allowing for people who are 60 feet tall!). This step may also look at simplifying data via dimensionality reduction techniques. Finally, this step often does some form of normalization and weighting in order to ensure that the further processing steps are acting on the right data. In the academic and commercial world, there are a number of technologies such as Data Civilizer [31], DataXFormer [32], Data Tamer [33], and Data Wrangler [34] that provide semi-automatic techniques to simplify data curation.

### 2.2.1 Anomaly and Outlier Detection

A particular task in data curation is the process of converting data, often with noisy inputs, to a version of the data that is amenable to further processing. The overall goal is to improve the signal-to-noise ratio such that further algorithms are likely to learn from the correct components of the data. The source of these errors can be due to a variety of factors such as sensor error, human error, etc. Although this is a very wide field in which significant human expertise is often used, one can also make use of machine-learning techniques such as unsupervised learning to automate or simplify the process. For example, a human analyst may set rules that are used on the data to ensure sufficient quality (such as height must be less than eight but greater than zero feet).

As an example of a more automated technique, given noisy data from a sensor, one may cluster the various data points into a set of clusters. Outliers from these clusters may be data points that are likely to be important—either because they are anomalous or otherwise. As noted in [35], there are three general approaches to outlier detection: 1) leverage unsupervised learning to look for outliers such as in the example above, 2) leverage a set of labels that correspond to normal or abnormal data in order to look for outliers, and 3) model only normal behavior with the intention of looking for samples that do not fit within the bounds of normal behavior.

One technique for anomaly and outlier detection can include techniques such as dimensional analysis [36], in which a spectral fingerprint of a dataset is created and outliers can be observed by looking for deviations from the expected fingerprint. Other techniques such as in [37] attempt to determine a background statistical distribution for a given dataset with the intention of quickly extracting components of a dataset that do not conform to the background distribution model.

The task of anomaly and outlier detection is a very important step and is often the limiting factor in quality of algorithmic performance in further processing steps.

### 2.2.2 Dimensionality Reduction

Often in a dataset, it is necessary to condense the amount of information that will be processed by the subsequent pipeline. This can be done for a variety of reasons such as improved computational performance, removal of redundant dimensions, or removal of features (dimensions) that will not play a part in further processing. In an image, for example, dimensionality reduction could be converting a color image with three channels to a single channel grayscale version that maintains important features of the original image such as edges or shapes without the additional red, green, and yellow channel information.

Techniques for dimensionality reduction look for variance in features within a dataset along with correlations across features. Through algorithms such as principal component analysis, users can quickly determine which features (or dimensions) of their dataset have the greatest variance and look for features that are closely correlated with other features. Using this information, it may be possible to keep only high-variance features and remove features that are closely correlated with other features. Thus, it may be possible to remove a large set of features within a dataset without adversely affecting future algorithmic performance.

### 2.2.3 Data Weighting and Normalization

Another technique for data conditioning is often referred to as data weighting or normalization. This process involves bringing various features, or dimensions, within a dataset to a common frame of reference or common dynamic range. If particular features have very high rates of change and/or very high dynamic range, it is possible that further processing steps will tend to overweight these changes compared to other features that may not have as high a range. Alternatively, there may be certain features in the dataset that should, in fact, have an outsized role in determining the output. In such cases, one may weigh these features higher than other features.

## 2.3 DATA LABELING

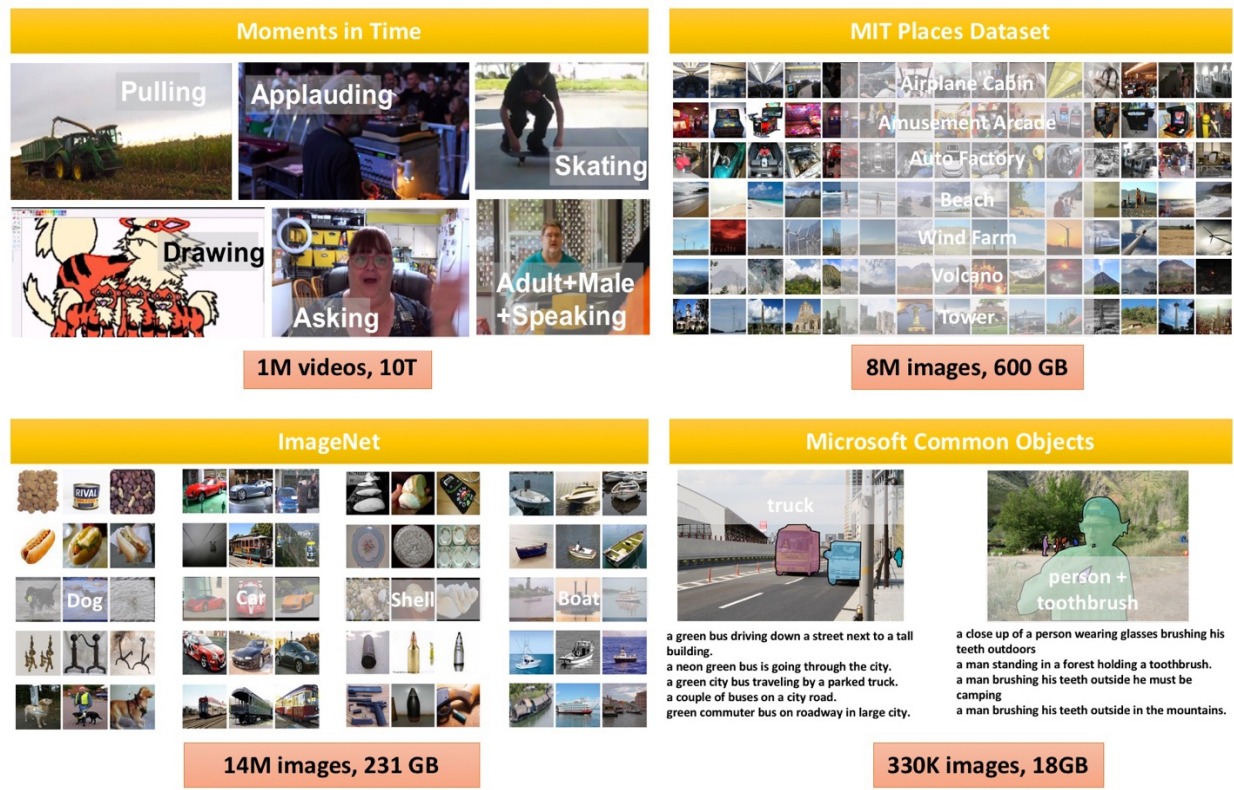

*Figure 2.4. Examples of popular curated datasets ImageNet, COCO, MIT Places, Moments in Time.*

One of the most time-consuming tasks in applying supervised learning techniques (such as neural networks) is in providing useful labels in the data. This task, sometimes referred to as data annotation, aims to provide the machine-learning algorithm with a set of labels that can be used to train the classification or regression model. Essentially, in supervised learning, the machine learns the pattern that associates an input with an output label. A multitude of labeled data points can provide a robust and repeatable training process. Of course, data collected from a sensor is rarely given clear labels and there has been a great deal of effort in the wider community to come up with techniques that can be used to simplify data labeling.

One obvious method to label a large quantity of data is to use human evaluators. A human evaluator (sometimes referred to as an "oracle") is often considered as the gold standard for labeling. While a developer of an algorithm or AI application can label a few images, this may not scale to thousands or millions of data samples. Further, a single evaluator can introduce bias (perhaps the evaluator has difficulty judging particular colors that may bias their labels). Figure 3.6 describes a few well-curated labeled datasets that are used widely in the community. For example, ImageNet [38] is a dataset that has helped spur the growth of recent computer vision advances. Other datasets such as MIT Places [39], Microsoft COCO [40], and the previously mentioned Moments in Time are also widely used by AI researchers.

One common technique for manual data labeling used across the commercial world is to leverage a service such as Amazon's Mechanical Turk [41, 42]. Using such a service, a user can upload a dataset of interest and quickly leverage a pool of users who can provide labels for the dataset. Although popular for tasks such as labeling of images or transcription of spoken languages, such crowdsourcing techniques often suffer quality issues [43] or do not scale to sensitive or proprietary datasets that require significant domain expertise such as those used across the DoD/IC. Within the DoD, one example of "crowdsourced" data labeling is being done via the Air Force's Project MAVEN [44]. This project aims to leverage the domain expertise of service members to label data where current techniques fail or cannot be applied due to the sensitivity of data.

Beyond manual annotation of data, the research community has been actively looking at techniques that provide varying levels of automation in the data labeling process by leveraging a human selectively or by leveraging machine intelligence on a small set of labeled data points.

In general, there are a number of algorithmic techniques that can still be used in cases where labels for data do not exist. For example, in the semi-supervised learning paradigm, a small subset of labeled data can be used in conjunction with unlabeled images. These semi-supervised techniques essentially infer the labels on data using a number of different techniques. For example, generative techniques assume that the subset of labeled data consists of labels for all classes of importance and attempts to learn statistical patterns of the labeled data. Then, assuming the unlabeled data looks similar to the labeled samples, it is possible to deconstruct the unlabeled data into the statistical components and compare with the labeled data. One particular instance is in the cluster and learn paradigm where the unlabeled data is clustered along with the labeled data. Unlabeled data is then assigned a label based on its proximity to a labeled sample. Another technique, referred to as self-training, may instead attempt to create a classifier using the labeled datasets and then iteratively apply the classifier to the unlabeled data. After application, instances where the classifier was very confident on the classification can now be added back to the pool of labeled data and the process can be repeated. Other techniques may use graph-based methods to represent the dataset where both labeled and unlabeled datasets form parts of the graph. Then, one can use graph matching techniques to apply labels to particular instances.

Another expanding area of research is in active learning. In this area, one attempts the above techniques as possible, but also makes use of a human observer to help the system. These techniques essentially allow both humans and machines to work together, with the machine applying labels to straightforward samples and focusing human attention on difficult samples.

Very generally, there are a number of ways in which one could still perform supervised learning in cases where labels are either nonexistent, limited, or difficult to collect:

1. **Semi-supervised learning**. In this paradigm, an algorithm is designed to leverage some labeled and some unlabeled data. For example, for a DoD/IC specific use-case, one may leverage a different but related dataset (using publicly available datasets on cars [45] to train models that can look for military vehicles).

2. **Active learning**. In this paradigm, a user is in the loop of the data preprocessing and labeling. In this technique, the system can apply labels to particular observations in which the confidence is high and can also leverage a human oracle for difficult or previously unseen samples.

3. **Self-supervised learning**. In this paradigm, the system does not need any explicit labels on the data samples to be provided. Rather than use labels, the algorithm can automatically extract metadata, encoded domain knowledge or other such correlations that can be used in place of explicit labels. For example, in [46], the authors learn the task of image colorization as a means to learn tracking.

4. **Automated labeling**. This technique leverages statistical techniques to apply labels to data. For example, in [47], the authors propose a technique using conditional random fields to build models that can be used to label data sequences.

## 2.4 DATA CONDITIONING WITHIN THE CONTEXT OF EXEMPLARY APPLICATION

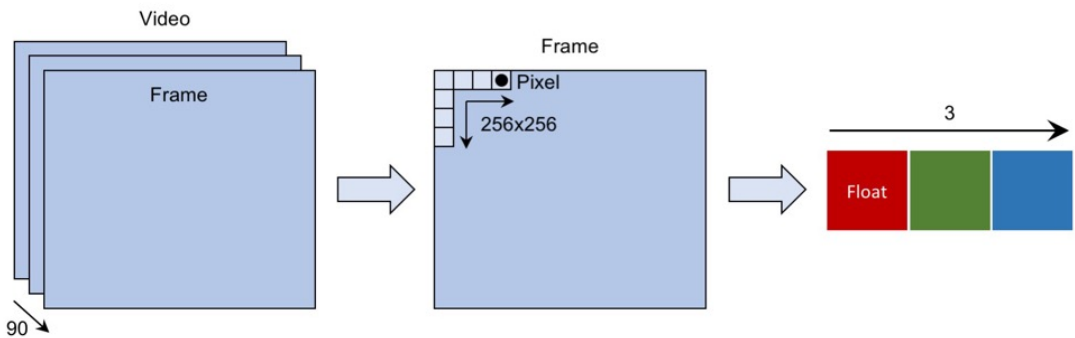

*Figure 2.5. Data conditioning within the context of video classification example. Data starts as raw compressed .mp4 videos, are converted to individual frames, and then resized to consistent dimension arrays across RGB channels.*

Although the Moments in Time Dataset is a well curated dataset with high quality labels, there are still a number of preprocessing steps that need to be taken. In developing our pipeline, the first thing we need to do is convert the MPEG-4 encoded video files (.mp4) to arrays that are easy for future processing steps. In the dataset provided, each video is 3 seconds long and consists of 30 frames per second (for a total of 90 frames per video). Each frame is 256×256 pixels and has three floating point values that correspond to the red, green, and blue (RGB) channels. Therefore, for each video represented in tensor format, this corresponds to a shape of 90×256×256×3 for a total of approximately 17.5 million integers per video. For 1 million videos, this corresponds to approximately 70 terabytes of raw video data. As a next step, we resize all of the frames to be a specific size: 224×224 pixels. Finally, we normalize the data so that all frames have a similar distribution of pixel intensities. First, we divide by 255 (the maximum range of pixel intensities) from each pixel intensity from each frame. Then we find the mean pixel intensity across all video frames and subtract this from the previous value. Next, we divide each pixel intensity by the standard deviation of pixel intensities. As a final step, in order to remove potential outlier frames, we remove all frames that contain pixel intensities below or above a certain threshold (in order to get rid of frames that are all black or white). This process is described in Figure 2.5.

At the end of this step, the raw videos are now cleaned up in a consistent format and stored on the file system as arrays that can be read in for further processing.

# 3  ALGORITHMS

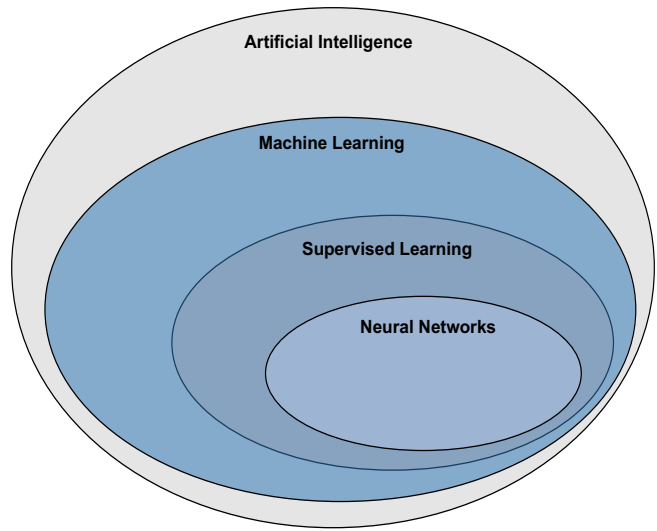

*Figure 3.1. Relationship between AI, machine learning, and neural networks. Figure adapted from Deep Learning by Ian Goodfellow.*

In the past decade, much of the hype around AI has come from advances in performance of machine-learning algorithms applied to various problems such as image classification. In this section, we will describe some of the popular machine-learning developments while highlighting a few salient algorithms.

From our perspective, machine-learning algorithms form the core of AI algorithms (with a few exceptions such as with expert systems). Figure 3.1, adapted from [48] describes, at a high level, the relationship between AI, machine learning, supervised learning and neural networks. While a lot of recent focus has been on neural networks, it is important to understand that there a multitude of machine-learning algorithms beyond neural networks that are used widely in AI applications. It is also interesting to note that many current and historical AI systems such as [49] actually leveraged techniques outside of traditional machine learning such as expert systems [50, 51] or the more general *knowledge-based systems* [52]. Knowledge-based systems leverage a knowledge base of information and an inference engine to apply the knowledge base to a particular application. Expert systems, a form of knowledge-based systems, utilize human experts to formulate a knowledge base that can be applied via an inference engine for decision making. Knowledge-based and expert systems continue to be used in a variety of applications such as tax software and were even included in early autonomous vehicles [53] [54]. For domains in which collection of data is limited, rules are complex and there is significant human expertise, expert systems can still play a major role in building an end-to-end AI system. Further, knowledge-based systems are often inherently explainable and interpretable, which can make them amenable to applications in which trust is critical.

At a high level, one can think of a tradeoff that exists between the ease of codifying human knowledge, compute power, and data availability. In cases with significant codified knowledge, limited compute power, and limited data availability, knowledge-based systems can play an important role in AI systems. Although the majority of our discussion is on machine-learning algorithms, knowledge- and expert-based systems are an important class of algorithms that still have wide applicability to DoD/IC applications.

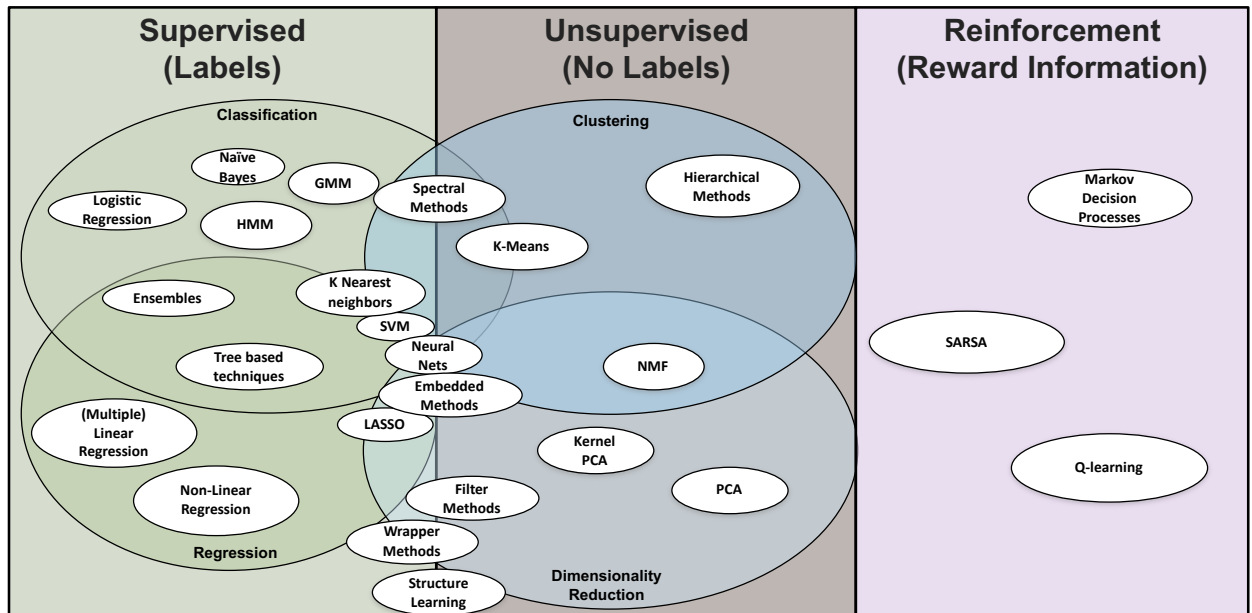

*Figure 3.2. Popular machine-learning algorithms and taxonomy. We acknowledge Lauren Milechin (MIT EAPS) and Julie Mullen (MIT LLSC) for contributing to the development of this figure.*

With the ability to collect large quantities of data, coupled with greater computational resources, the shift in technology has been toward the world of machine learning. In the machine-learning paradigm, a user provides a system with (varying amounts of) data along with a set of rudimentary guidelines (such as generative model, number of classes, etc.). Using this input, the machine "learns" a mapping that can be applied to the data in order to generate the required output. With this paradigm, machine-learning algorithms essentially learn the relationship between inputs and outputs.

To highlight the many techniques for machine learning, we use the taxonomy presented in Figure 3.2. Other taxonomies, such as those from [55], are alternate ways of organizing the variety of machine-learning algorithms. In our taxonomy, machine-learning algorithms are broadly broken into supervised, unsupervised, and reinforcement learning techniques. Supervised learning techniques have labels that relate inputs and outputs; unsupervised techniques typically do not; and reinforcement learning algorithms are provided with reward signals rather than explicit labels. Although this is a good first-order break-up of the landscape, we would like to note that these boundaries are not meant to be rigid and in practice, there are a number of algorithms that cross these boundaries. In the remainder of this section, we provide a brief overview of each of these learning paradigms along with a discussion of where these techniques may be used in the development of AI systems.

## 3.1 SUPERVISED LEARNING

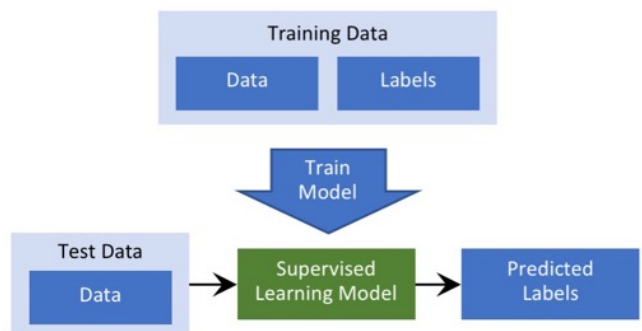

*Figure 3.3. Supervised learning makes use of data and labels in order to train a model that automates this process.*

Supervised algorithms start with labeled data (or ground truth) and aim to build a model that can be used to predict labels for data in which labels (or classifications) do not exist. As shown in Figure 3.3, supervised learning makes use of data and labels to train a model that can be applied on test data in order to predict future labels. Thus, the video classification task we present is a supervised learning problem because we are given labels of the video samples that are used to train a model that we use to classify new data samples. Generally, supervised learning algorithms attempt one of the following goals: regression (to predict a future continuous variable) or classification (to predict a new class or label). In most cases, the majority of computing time in supervised learning is spent training a model from training examples. The model generated by the algorithm is essentially the representation of what the system believes relates the inputs to the corresponding outputs. After training, the model can now be applied for inference tasks. During inference, the trained model is applied to previously unseen test data to predict labels in the case of classification or future values in the case of regression.

To further explain supervised learning, consider the versatile $k$-nearest neighbors algorithm [56]. This algorithm can be applied to both classification and regression problems and relies on the assumption that similar points (data points that are close to each other in a feature space) have similar labels/outputs. To use this algorithm, each data point in a dataset is represented in a multi-dimensional feature space that corresponds to data features (for example, in an image, this could be the pixel intensities and/or pixel locations) that represent each individual data point. With this representation, it is possible to predict values or apply labels to new data points. For a new data point represented in the same feature space, we find the $k$-closest neighbors (i.e., we look for all data points that are close by in the multi-dimensional space). If performing regression, the predicted value is an average of the neighboring values. For classification, the $k$-nearest neighbors vote with their label and the predicted class label for the new data point is the label held by a majority of the $k$-nearest neighbors. Although this algorithm is very easy to implement and has been used in a variety of applications, it can be sensitive to too many features (or dimensions) or in feature spaces where the closeness assumption or definition does not hold. Figure 3.4 describes an example of applying the $k$-nearest neighbors algorithm to a two-dimensional dataset. On the left side of the figure, we describe how one would classify a data point when we have labels corresponding to three labels. The new data point in this figure is represented by the star-shaped dot. The $k$=7 nearest neighbors are highlighted with circles. On the right side, we describe how one could use the same algorithm for regression. In this example, we are trying to predict the value of the second dimension. Similar to the classification example, for a new data point, we find the *k*=7 closest neighbors and the predicted second dimension value is given as an average of the second-dimension value of the seven nearest neighbors.

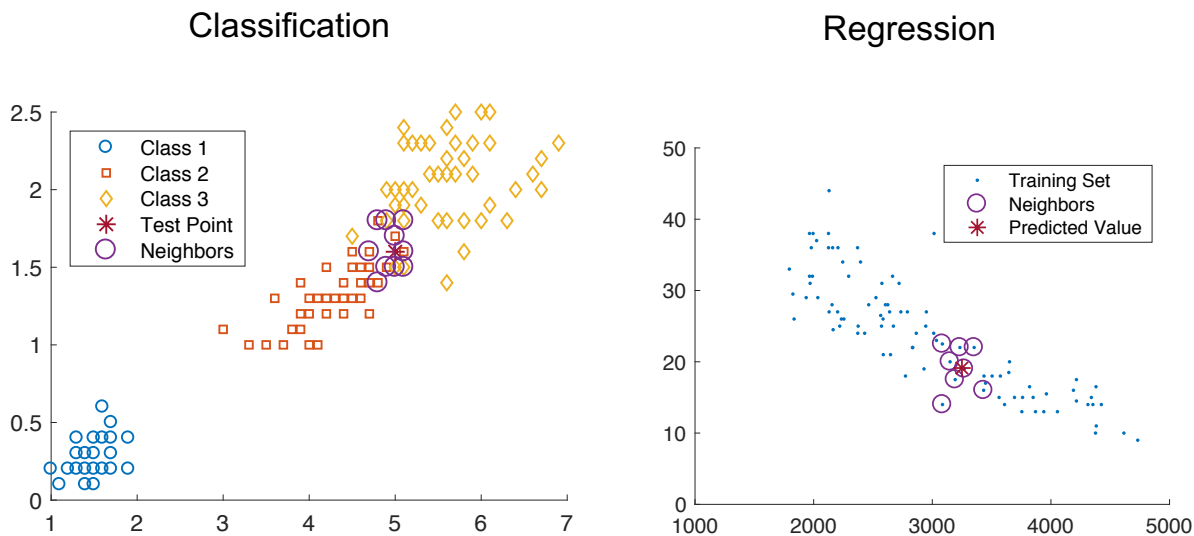

*Figure 3.4. Using a k-means supervised learning technique for classification and regression.*

Supervised learning algorithms play an outsized role in many of the recent machine-learning advances. The quality of supervised learning algorithms, however, largely relies on access to high-quality labeled data. Although there are a number of labeled datasets that have yielded breakthrough advances in fields such as image classification, voice recognition, game play, and regression, in DoD/IC applications, these may not be available. In such cases, unsupervised learning provides a means to analyze data.

**3.2 UNSUPERVISED LEARNING**

Unsupervised learning is a technique that applies statistical analysis techniques to data in which explicit labels are not provided. Very often, upfront data conditioning is done via unsupervised learning with the aim of looking for outliers or reducing the dimensionality of data. Without data labels, it is difficult to classify data points, and unsupervised learning techniques are limited to clustering or dimensionality reduction. More formally, if we observe data points $X_1, X_2, X_3,…,X_N$, we are interested in looking for patterns that may occur among these data points. In this example, each data point X can be made up of a number of features or dimensions. For example, in an image, each data point X may correspond to a single pixel and each data point can be represented by three features—the RGB value associated with a single pixel.

In unsupervised learning, there are often no clear metrics such as accuracy or recall for the algorithm. However, given that unsupervised learning algorithms can work on unlabeled data, they are often an important first step in any AI pipeline.

Typically, unsupervised learning is used for clustering and data projection. Clustering algorithms group objects or sets of features such that objects within a cluster are more "similar" than those across clusters. The definition of similarity is often highly dependent on the application. For example, in an image processing application, similarity may imply the difference in pixel intensities across different image channels; in other applications, similarity may be defined as Euclidean distance. Measuring intra-cluster similarity is often done via measures such as squared error (se):

$$se = \sum_{i=1}^{k} \sum_{p \in c_i} \|p - m_i\|^2$$

Many clustering algorithms work by iteratively trying to minimize error terms, such as the squared error term defined above, by placing different data points in different clusters and measuring the error. Within clustering, common algorithms include *k*-means, nearest neighbor search, spectral clustering. Figure 3.5 shows a pictorial example of a notional clustering algorithm applied to data within a two-dimensional space. For the purpose of this figure, the "similarity" is defined to be Euclidean distance (the geometric distance between two points).

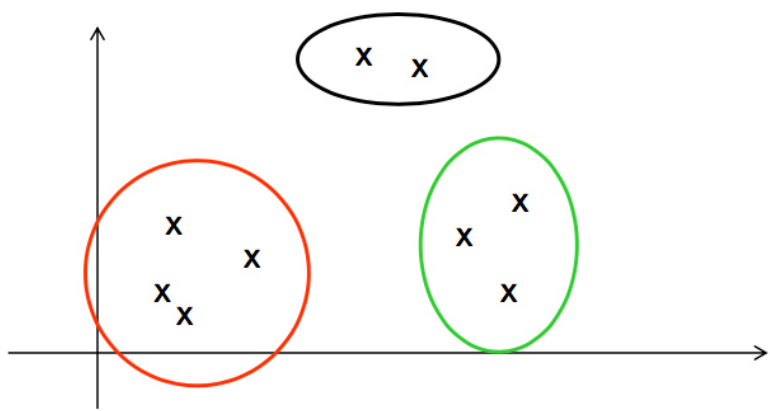

*Figure 3.5. Notional clustering example for data represented by two features.*

Within data projection/preprocessing, typical tasks include principal component analysis (PCA), dimensionality reduction, and scaling. Dimensionality reduction is used to reduce a large dataset into a more efficient representation comprised of high variance dimensions. These techniques can be especially useful to simplify computation or represent a dataset. Typically, one uses these techniques for selecting a subset of important features or representing data in a lower number of dimensions.

Consider the example of using PCA for dimensionality reduction. In this technique, a set of possibly correlated variables are converted to a set of uncorrelated variables using orthogonal transformations. In essence, through this technique, we are looking for a lower dimensional space representation of a dataset that still maintains some of the broad trends in the data.

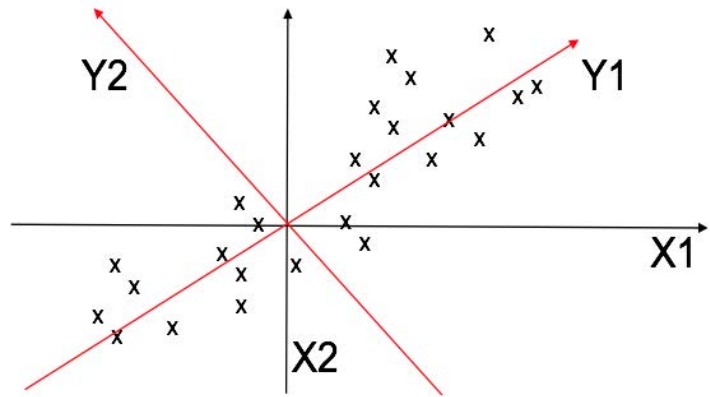

*Figure 3.6. Principal component analysis. In this example, Y1 is the first principal component and Y2 is the second principal component.*

For example, Figure 3.6 describes an example two-dimensional dataset that was originally represented by the axes X1 and X2. After applying PCA, the axes Y1 and Y2 are found to be the principal components. Using this new representation, the PCA technique says that if you would like a lower dimensional (in this case, one-dimension) representation of the dataset, you can convert it to the axes of Y1. As can be seen in the Figure 3.6, axis Y1 does a good job of representing the spread in the data and would be the best 1D representation of the data.

## 3.3 SUPERVISED AND UNSUPERVISED NEURAL NETWORKS

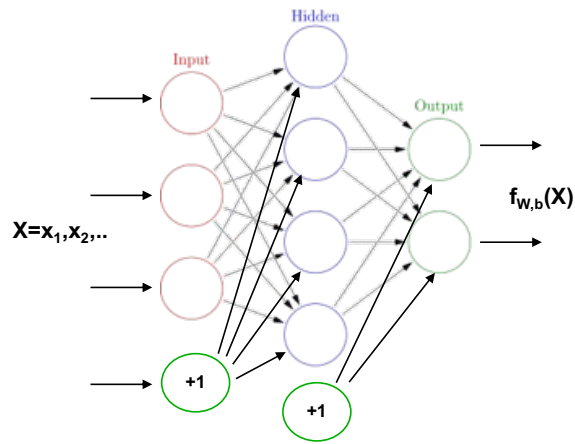

*Figure 3.7. Notional neural network.*

Neural networks are a popular machine-learning technique that are largely used for supervised learning but can be applied to unsupervised learning problems as well. The biologically inspired computing systems learn by repetitive training to do tasks based on examples (training data). A neural network consists of inputs, layers, outputs, weights, and biases. Deep neural networks (DNNs) differ from traditional neural networks by having a large number of hidden layers. DNNs have had much success in the past decade in a variety of applications and are supported by a number of toolboxes [57–59] and hardware platforms. Popular extensions of the neural network computation model include convolutional neural networks [7], recursive neural networks [60], and deep belief networks [61]. The network shown in Figure 3.7 in an example of a feedforward neural network that consists of one input layer, one hidden layer and one output later. The arrows in the figure indicate connections across neurons.

In general, for a neural network of *k* layers, the following expression describes the input-output behavior between any two layers (*l* and *l+1*):

$$y_{l+1} = f(W_l y_l + b_l)$$

where $y_l$ corresponds to the outputs of layer *l*, $W_l$ the weights between layers *l* and *l+1* and $b_l$ the corresponding biases. In this equation, the function *f(.)* corresponds to a non-linear activation function such as the sigmoid, rectified linear unit or tanh function [62]. During the training phase of a neural network algorithm, the weights *W* and biases *b* are iteratively adjusted in order to represent the often non-linear relationship between inputs and outputs. The training phase consists of the following steps:

1. **Forward Propagation:** A training example (or set of examples) is fed through the network with some initialized weights and biases. The output is computed and an error term is calculated as the difference between the computed output(s) and the intended (ground truth) output(s).

2. **Backward Propagation:** Given the error from the previous step, network weights are adjusted to better predict the labels of future unlabeled examples. This is done by adjusting each weight in proportion to its individual contribution to the overall error. As the contribution to the error from weights in earlier layers depends on the contribution from weights in later layers, the error signal is described as "backward propagating" through the network. Iterative optimization techniques built on this idea, such as the popular stochastic gradient descent (SGD), are thus known collectively as "back propagation."

The above procedure is repeated for as many training examples as possible and when a particular convergence criterion (such as accuracy) is met, the model is considered trained and can be used for inference. In practice, it is common to split the total training set into training and validation subsets. This can help reduce the likelihood of overfitting in which the trained model works especially well on the training data but does not generalize to new unseen samples. Typically, the training phase is where the majority of computation takes place. Once a model is trained and validated, it is ready to be used for inference.

Inference is the phase in which a trained model is applied to new data in order to predict new values (i.e., a forward pass through the network with weights and biases set by the training step). In this step, the model computes an output for a previously unseen input. If the model was trained well, the output should correspond to the correct output for a given input example.

Although many uses of neural networks focus on supervised learning problems, there are many ways to use neural networks for unsupervised learning. Network architectures such as autoencoders or self-organizing maps are often used as techniques to use neural networks for unsupervised learning problems. Another technique, generative adversarial networks (GANs) [63] (not to be confused with adversarial AI, discussed later in this section), uses two neural networks—a generator and discriminator—to train a model. In this architecture, the generator creates artificial samples meant to represent data from the training dataset.

These artificial samples are then passed to the discriminator, which decides whether the sample is real or fake. Continuing this process iteratively allows the generator to improve the quality of its artificial samples and the discriminator to improve its discriminative capabilities. GANs can be an important tool in understanding the structure and patterns present in unlabeled datasets.

At present, the theory behind DNNs—why they work, how to estimate performance, performance bounds, etc.—is not well understood and is an active area of fundamental research. A user wishing to develop a neural network solution will typically be faced with a number of choices such as model architecture (what type of network), number of layers, activation functions, learning rate, batch size (for memory limited applications), etc. Although there are some high-level guidelines that can be used by practitioners developing networks for their application, much of the state-of-the-art relies on exploration of parameters. There are also techniques such as in [64] that can leverage high performance computing to look at this vast parameter space more efficiently, but much of the current state-of-the-art relies heavily on domain knowledge and experience. Understanding the theory behind DNNs is particularly important for DoD and IC applications where we will need to adopt solutions from other applications. At present, without a clear understanding of the theory behind DNNs, this process is largely ad-hoc and relies on costly trial-and-error solutions.

One generally accepted finding is the notion that deeper neural networks (more layers) typically perform better. This is likely due to the fact that more layers allow for the network to provide decision boundaries that are more non-linear and much more complex decisions. Current state of the art networks such as ResNet and Inception [65] often consist of hundreds of layers. As the community looks toward even deeper networks ($10^5$–$10^6$ layers) it is likely that we will see the emergence of a new type of neural networks—sparse neural networks [66] [67]. As opposed to the commonly used dense equivalent, sparse DNNs will consist of a large number of weights that are zero. This can be quite advantageous for hardware platforms where memory or computations are resource constrained.

### 3.4 TRANSFER LEARNING

As described in [43], traditional machine-learning techniques assume that the data used to train and deploy a model are of the same domain. The framework of transfer learning allows the training of models in a source domain or feature space and the deployment of this model to a target domain in which sufficient training data may not be available and domain or feature space are different. Transfer learning can happen in different settings. The first, inductive transfer learning, occurs in cases where there are labels in the source and target domain; on the other hand, transductive transfer learning is used in cases where only source labels are available but target labels are not (similar to the unsupervised learning case), and unsupervised transfer learning is a case in which neither source nor target labels are available [43].

Transfer learning may be of particular interest to DoD/IC applications where training data within domains or tasks of interest is scarce. Thus, within the framework of transfer learning, it may be possible to find domains or tasks that are related to the target domain or task with abundant training information. Then, using inductive, transductive, or unsupervised transfer learning techniques, it may be possible to transfer the knowledge gained from the source domain and task to the target domain and task with significantly less effort than trying to train a model for the target domain and task from scratch. Although a promising avenue for applications such as those in the DoD and IC where we are data rich but truth (labeled) poor, a theoretical understanding of how well transfer learning works for arbitrary applications is still limited.

## 3.5 SEMI-SUPERVISED LEARNING

In many cases, one has access to a small set of labeled data but wishes to make additional use of large quantity of unlabeled data to improve the training of their models. Semi-supervised learning algorithms are a class of algorithms designed to work within this paradigm. The authors of [68] provide a good overview of the variety of techniques that fall within semi-supervised learning. Semi-supervised learning differs from supervised learning, which works on all labeled data and unsupervised learning in which none of the data is labeled. Further, it should be noted that semi-supervised learning has certain assumptions about the quality and relationship between labeled and unlabeled samples. For example, it is assumed that the labeled samples provide coverage over the possible classes and are related in some feature space to unlabeled samples that should have the same label. Thus, it is an important paradigm, but simply having access to a dataset with labeled and unlabeled samples is not necessarily sufficient to implementing a semi-supervised algorithm. As noted in [68], a few of the many ways in which semi-supervised learning can be used include the following.

1. **Generative models.** In this case, it is assumed that the dataset can be well represented by some sort of statistical mixture model. Then, if one has at least one labeled sample for each mixture component, it is possible to leverage the unlabeled samples to improve model quality. This process can also be extended to use any clustering algorithm to apply labels to the unlabeled samples.

2. **Self-training**. In this case, we use the labeled samples to create a classifier that can be then used to classify the unlabeled samples. High-confidence classifications are then added to the labeled sample pool and a new classifier is trained. This process is repeated until satisfactory results are achieved

3. **Graph-based**. In this class of semi-supervised learning algorithms, data points (both labeled and unlabeled) are represented as a graph. Using techniques such as graph similarity, it is possible to infer labels on the unlabeled samples.

Although at first glance it seems that semi-supervised learning may be a natural fit to DoD and IC problems by greatly reducing the burden on labeling samples, it should be noted that the use of unlabeled samples in conjunction with labeled samples provides no guarantee of improving supervised learning techniques on the labeled samples alone. Further, using semi-supervised learning typically involves significant manual design of features and models in order to achieve meaningful results [69].

## 3.6 REINFORCEMENT LEARNING

Reinforcement learning is another machine-learning paradigm that has received significant interest in the recent past [70]. Major breakthroughs of reinforcement learning can be seen in robotics [71], learning to play Atari games [72], and improving computer performance in the game of Go [73].

In contrast to other learning paradigms, reinforcement learning leverages a reward signal in order to learn a model. Thus, this reward signal can provide much higher level "labels" that the system can eventually use to learn from and can work particularly well in complex environments where it may be difficult to tease out specific rules that need to be learned. An example of reinforcement learning in action is given in [74]. In this work, researchers use reinforcement learning techniques to develop an algorithm capable of controlling a helicopter. As the authors note, the difficulty in modeling the physical properties of a helicopter make it a good candidate for reinforcement learning.

There are a number of factors that make reinforcement learning different from other learning techniques. First, within reinforcement learning, there is no supervisor but only a reward signal. Second, the reward signal or feedback is not instantaneous but often delayed. Within reinforcement learning, signals are often provided sequentially and time or order of samples and signals are important. At each step, the response to subsequent samples differ. For the example of using reinforcement learning to learn to play Atari games, at the beginning, the rules of the game are unknown and the system learns directly from interactive game play. The system picks an action on the joystick and sees a set of pixels and scores that correspond to a positive or negative signal that is used to adjust behavior.

## 3.7 COMMON ALGORITHMIC PITFALLS

When designing machine learning algorithms, developers should be aware of a number of common errors that can arise. Below are a few issues that we have observed in practice.

- **Over-fitting ("variance") vs. under-fitting ("bias").** Over-fitting is a phenomenon when a particular machine-learning model is too closely fit to the training examples. When over-fit, an algorithm may have trouble generalizing to new examples. Under-fitting, on the other hand, is when an algorithm provides an overly simplistic model that describes the training examples. There are a number of ways that one can avoid such issues—for example, selecting training data that represents the variety of examples to be seen, including a regularization term in the training objective, or choosing different algorithms.

- **Bad/noisy/missing data**. In this challenge, a model is trained on bad, noisy, or missing data. In such cases, the trained model may not work as intended and decisions may be made on the wrong set of features. In the case of missing data, the model may ignore important features or make certain assumptions of the data that are unlikely to work in practice. Overcoming this challenge often requires the use of good data conditioning techniques of human intervention to ensure the fidelity of training data.

- **Model selection**. It is imperative that the model being used to represent the desired input-output relationship be closely related to the actual input-output relationship.

- **Lack of success metrics.** Having a clear metric of algorithmic success is important. Although metrics such as accuracy or precision provide some view into the performance of the algorithm, there may be other metrics that should be carefully designed prior to training a model.

- **Linear vs. non-linear models**. Picking the right type of model is also important. If there is to be a linear relationship between inputs and outputs, one should pick a linear model.

- **Training vs. testing data**. Carefully segregating data into training, validation, and testing datasets is an important best practice that can help avoid issues such as over-fitting or under-fitting.

- **Computational complexity, curse of dimensionality**. Data conditioning techniques can be used to help reduce the dimensionality of datasets which can also help machine-learning algorithms use cleaner data when determining input-output relationships.

## 3.8 ALGORITHMS WITHIN THE CONTEXT OF EXEMPLARY APPLICATION

For the exemplary video classification example, we leverage a number of open-source models. In order to greatly reduce training time on the very large array, we use pre-existing models and weights and update them in a process similar to that outlined in the transfer learning section.

To begin the process, we tested many different types of models in order to judge the model that is likely to work well for the classification task at hand. Specifically, we focused on the following considerations:

1. Model type—spatial, temporal, relational or auditory

2. Training from scratch vs using pretrained models

3. Types of input channels

4. Computational constraints such as memory and performance

To simplify our problem, we considered each video as a series of images. Using this simplification, we were better able to focus on spatial models such as convolutional neural networks. Given the rich variety of preexisting models for image classification, we decided to use transfer learning from existing models such as Inception and ResNet. Given the simplification of videos to series of images, we only used the RGB channels of the video and ignored the audio tracks. Each node on our system consists of NVIDIA K80 graphics processing units (GPUs) with 16GB of memory, which is not enough to load the full training dataset. Thus, we developed a batch training mechanism that could iteratively train on a smaller subset of training samples rather than the full set. Determining other parameters such as number of layers, learning rate, and batch size was done by using HPC techniques in order to quickly look at the thousands of possible parameter settings.

# 4    COMPUTING

Many recent advances in AI can be at least partly credited to advances in computing hardware [38, 75]. In particular, modern computing advances have been able to realize many computationally heavy machine-learning algorithms such as neural networks. While machine-learning algorithms such as neural networks have had a rich theoretic history [76], recent advances in computing have made the application of such algorithms a reality by providing the computational power needed to train and process massive quantities of data. While the computing landscape of the past decade has been rich with numerous innovations, DoD and IC applications that require covert and low size, weight, and power (SWaP) systems will need to look beyond the traditional architectures of central processing units (CPUs) and GPUs. For example, in commercial applications, it is common to offload data conditioning and algorithms to non-SWaP constrained platforms such HPC clusters or processing clouds. The DoD, on the other hand, may need AI applications to be performed inside low-SWaP platforms or local networks (edge computing) and without the use of the cloud due to insufficient security or communication infrastructure. Beyond modern computing platforms, the wide application of machine-learning algorithms has also been supported by the availability of open-source tools that greatly simplify developing new algorithms. In this section, we highlight some recent computing trends along with a brief introduction to software packages that have relevance to DoD and IC applications.

## 4.1    PROCESSING TECHNOLOGIES

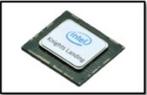

*Figure 4.1. Some examples of processing technologies with salient features.*

Figure 4.1 describes the primary classes of processing technologies for AI applications. Although CPUs continue to dominate in terms of availability, cost, and market support, many of the recent advances in machine-learning algorithms, specifically neural networks, have been driven by GPUs. GPUs are essentially parallel vector processing engine, which have shown themselves to be adept at massively parallel problems such as training neural networks and are particularly well-suited to the back propagation algorithm described in Section 3.3. While GPUs will likely dominate supervised learning model training in the near future, it is important to note that there are also a number of academic and commercial groups developing custom processors tuned for neural network inference and training. An example of such a processor is the Google TPU [75, 77], which is an application specific integrated circuit (ASIC) originally designed for machine-learning inference, while more recent versions support both inference and training [78]. There is also significant research in new hardware architectures such as neuromorphic computing [79], which may work well in low-power or resource-constrained environments.

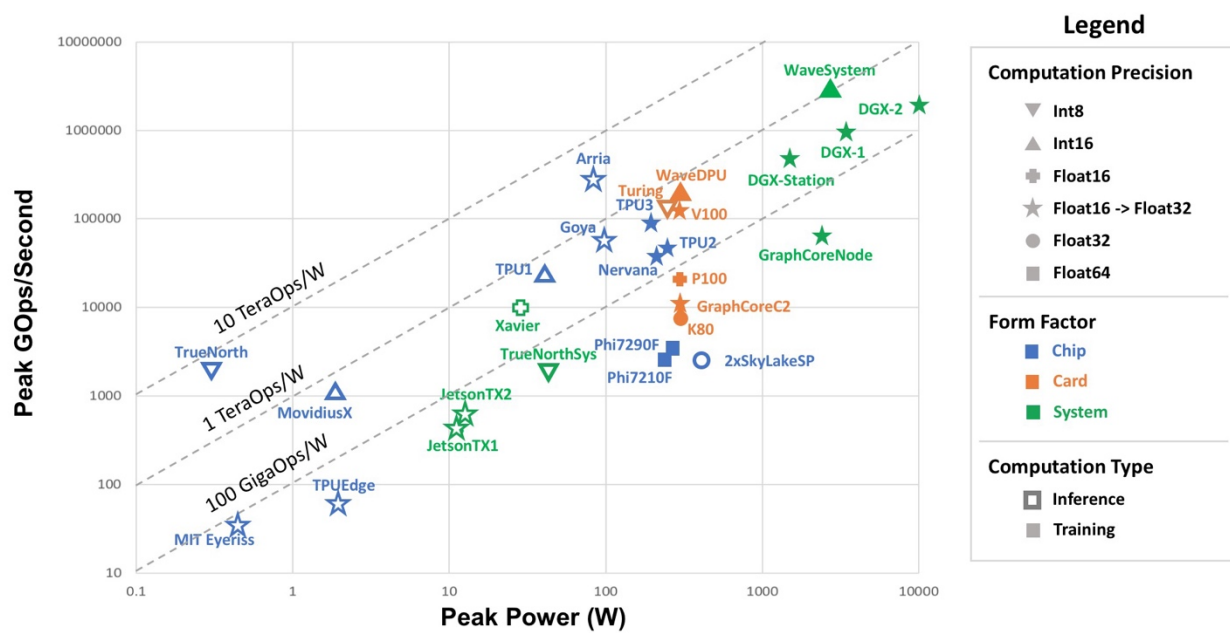

*Figure 4.2. December 2018 view of AI computing systems. The x-axis indicates peak power and the y-axis indicate peak giga-operations per second. (GOps/s) Note the legend on the right that indicates various parameters used to differentiate computing techniques. The computational precision of the processing capability is depicted by the geometric figure used. The processing spans from single byte Int8 to 8-byte Float64. The form factor is depicted by the color; this is important for showing how much power is consumed, but also how much computation can be packed onto a single chip, a single PCI card, and a full system. Blue is only the performance and power consumption of a single chip. Orange shows the performance and power of a card (note that they all are in the 200-300W zone). Green shows the performance and power of entire systems—in this case, single node desktop and server systems. We limit this here to single motherboard, single memory-space systems because it can get out of hand if a variety of multiple node systems are included. Finally, the hollow geometric figures are performance on inference only, while the solid geometric figures are performance on training (and inference) processing. Mostly, low-power solutions are only capable of inference, though there are some processors (WaveDPU, Goya, Arria, and Turing) that are targeting high performance on inference only.*

Figure 4.2 graphs some of the recent processor capabilities (as of December 2018) mapping peak performance vs. power usage. As shown in the figure, much of the recent efforts have focused on processors that are in the 10–300W range in terms of power utilization, since they are being designed and deployed as processing accelerators. (300W is the upper limit for a PCI-based accelerator card.) For this power envelope, the performance can vary depending on a variety of factors such as architecture, precision, and workload (training vs. inference). At present, CPUs and GPUs continue to dominate the computing landscape for most AI algorithms. However, the end of Moore's law [80] and Dennard scaling [81] implies that traditional commercial off-the-shelf (COTS) technologies are unlikely to scale at the rate at which computational requirements are scaling. For example, according to [82], the amount of computation required for training popular neural network models goes up at a rate of approximately 10×/year. To address these challenges, a number of hardware developers have begun to develop customized chips based on field-programmable gate array (FPGA) or application-specific integrated circuit (ASIC) technologies. For example, the aforementioned Google TPU [83], Wave Computing Dataflow Processing Unit (DPU) [84], GraphCore C2 [85], and Habana Goya [86] are all ASICs customized for tensor operations such as parallel multiplications and additions. Other ASICs have been designed to push the boundaries of low-power neural network inference, including the IBM TrueNorth neuromorphic spiking neural network chip [79] and the MIT Eyeriss architecture [87]. One interesting trend of note is that many hardware manufacturers, faced with limitations in fabrication processes, have been able to exploit the fact that machine-learning algorithms such as neural networks can perform well even when using limited or mixed precision [88, 89] representation of activation functions, weights, and biases. Such hardware platforms (often designed specifically for inference) may quantize weights and biases to half precision (16 bits) or even single bit representations in order to improve the number of operations/second without significant impact to model prediction, accuracy, or power utilization. For example, as reported by NVIDIA, the V100 GPU can perform 7.8 teraFLOPS of double-precision, 15.7 teraFLOPS of single-precision, and 125 teraFLOPS of mixed single- and half-precision [88].

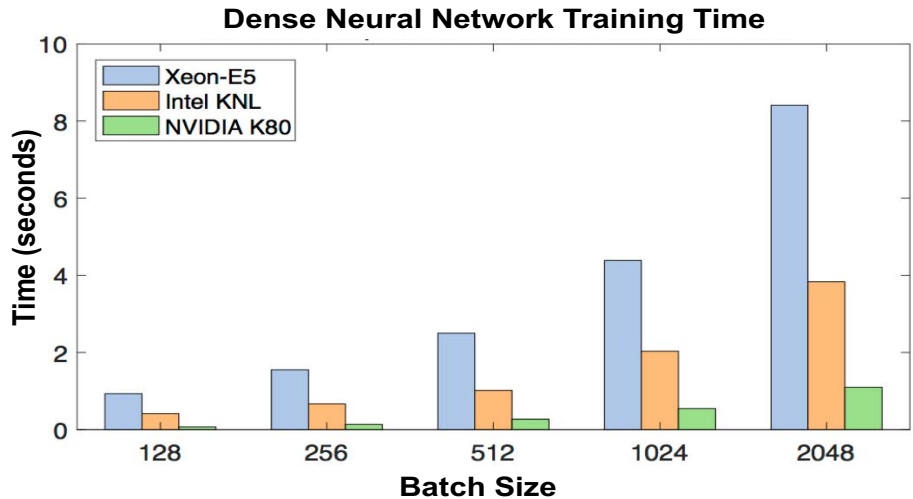

*Figure 4.3. Performance results for training convolutional neural network using different hardware platforms as a function of batch size. Evaluation was performed using TensorFlow and training the AlexNet model with data from the ImageNet dataset.*

Although looking at published numbers from vendors provides an important view of performance, these numbers may be derived from algorithms and applications that are particularly well suited to the hardware platform. In most cases, it is also valuable to benchmark performance by testing various hardware platforms on workloads or applications of interest. For example, Figure 4.3 shows the time taken to train a dense convolutional neural network model using different hardware platforms. As expected, the NVIDIA K80 GPU performs the best when compared against the time taken for training by the Intel Xeon-E5 and Intel Knights Landing processor.

Beyond performance, power utilization, or other resource constraints, choosing the right computing technology for an application may also be driven by the software being used and what hardware platforms are supported by this software.

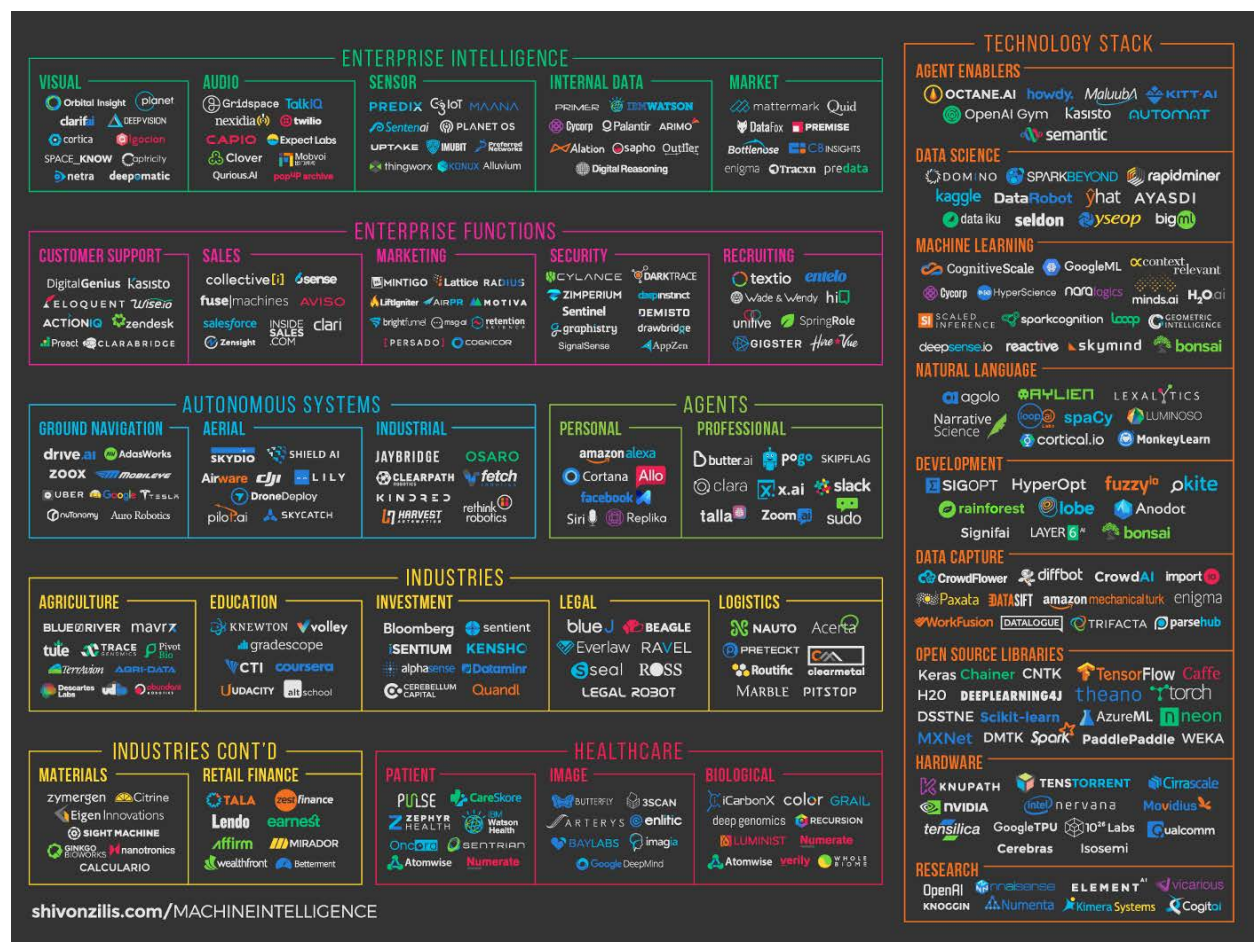

*Figure 4.4. Survey of commercial organizations developing AI software tools. This infographic was developed by Shivon Zilis.*

## 4.2 MACHINE LEARNING SOFTWARE

The field of available software for AI and machine learning has witnessed an explosion of options in the past few years. For example, Figure 4.4 shows some of the many tools available. These tools typically provide a high-level domain-specific interface that allows users to quickly apply AI and machine learning techniques to problems and domains of interest.

As noted in the previous section, choosing a software environment for development and hardware platform for deployment are not necessarily independent decisions. Particular environments may only work with a subset of hardware platforms (or at least work well with a subset). A recent industry trend has been for well-known machine-learning software providers to develop custom hardware tuned to work well for their software environment. For example, software packages such as TensorFlow, PyTorch, MXNet, RAPIDS, CNTK are led by Google, Facebook [57], Amazon [90], NVIDIA, and Microsoft [91], respectively. Each of these vendors has also publicized the fact that they are developing hardware platforms (most often focused on inference) that will provide benefits to the users of their software packages. These benefits could include inexpensive use of proprietary cloud-based computing solutions, higher performance or additional software functionality. When deciding which of the multitude of AI and machine-learning frameworks to use for an application, it is important to understand how they fit in to the larger computing pipeline. Further, it should be noted that most of these tools output models in a format unique to that software package. Fortunately, there are efforts such as the Open Neural Network Exchange (ONNX) to standardize model specifications and allow developers to move models from one ecosystem to another. However, as of December 2018, the portability of models across platforms is still limited. Finally, deploying individual software packages in private clouds or HPC clusters still requires significant efforts to reach published performance numbers.

### 4.3 HIGH-PERFORMANCE COMPUTING

HPC systems play an important role in developing AI systems. Often, HPC systems are used for data conditioning and algorithm development. In the realm of data conditioning, HPC systems are particularly well designed for processing massive datasets in parallel by providing high-level interfaces and high-quality hardware. In the realm of algorithm development, HPC systems can be used for model design and training computationally heavy models such as neural networks. The parallel processing capabilities and abundant storage present in most HPC systems make them particularly well-tuned to such computationally heavy tasks that may require large sweeps of model parameter spaces. However, there are some potential pitfalls with using HPC systems for developing AI systems including internode communication overhead, data distribution, and parallelizing computations [92]. HPC systems can be used to train and optimize AI algorithms and models by evaluating many parameters (parallel hyperparameter training) [93, 94] or by training single algorithms and models on many compute nodes of an HPC system [95–97].

Many modern HPC systems such as the Oak Ridge National Laboratory's Summit system [98] feature hybrid architectures consisting of heterogenous computing elements such as CPUs, GPUs, and FPGAs. Although HPC systems have a long history, the iterative nature of AI and machine learning development has led to new software and tools. In our research at MIT Lincoln Laboratory, we have spent significant effort in developing new tools [99, 100] that enable interactive, on-demand rapid prototyping capabilities for AI and machine-learning practitioners interested in applying the power of HPC systems to AI and machine-learning workloads.

Table 1 describes a number of salient differences between traditional HPC and AI/machine-learning application development and execution. The new MIT Lincoln Laboratory tools bridge these differences, and they enable both HPC and AI/machine-learning application development and execution on the same system and in the same user environments.

Table 1

**Salient Differences Between Designing High Performance Computing Systems for Traditional Workloads vs. Machine-Learning Workloads**

|  | **Traditional Workloads** | **AI/Machine-Learning Workloads** |
|---|---|---|
| **Programming Environments** | OpenMP [101], MPI, C, C++, Fortran | Tensorflow, Caffe, Python, MATLAB, Julia |
| **Software** | Designed for clusters | Designed for laptop |
| **Deployment** | Bare Metal | VMs/Containers |
| **Computing** | Homogenous | Heterogenous |
| **Computing Literacy** | High | Low |
| **Scheduling** | Batch | Interactive |
| **Resource Managers** | Slurm [102], SGE [103] | Mesos [104], YARN [105], Slurm |
| **Data Sources** | Generated by simulation | Imported from outside |
| **Data Storage** | Files | Databases (in-memory, file-based) [106] |
| **Interfaces** | Terminal | Jupyter/Web-based [100] |

## 4.4  COMPUTING WITHIN THE CONTEXT OF EXEMPLARY APPLICATION

For our particular application, we use Nvidia K80s as the computing platform. As described in Figure 4.3, the K80 GPU performs well when compared to other hardware platforms that were available on our system (Intel Xeon-E5 and Intel KNL/Xeon64c). Although GPUs can perform much of the model learning, we still utilize a CPU (Intel Xeon-E5) for data preprocessing and manipulation. The reason we use two different processors is three-fold: 1) our cluster has many more available CPUs that allow us to use parallel processing techniques for data conditioning, 2) each CPU has significantly more memory available (256 GB for a CPU vs. 16 GB for a GPU) that makes it amenable to manipulating large datasets, and 3) preprocessing is not amenable to vector-parallel processing. To fit datasets in the limited GPU memory for training, we used a batch training technique such as that utilized in [38] so that weight updates occur on a smaller batch of data when compared against the nearly one million videos. For the software environment, we used the widely available open-source TensorFlow and Keras [107] packages developed by Google that support the CPUs and GPUs available on our cluster. Development was performed on the MIT SuperCloud [108, 109] cluster at the Lincoln Laboratory Supercomputing Center. The MIT SuperCloud computing environment provides access to a variety of hardware and software platforms necessary for rapid prototyping of AI solutions [110].

Figure 4.5 describes the computational performance of various steps in the machine-learning pipeline: 1) loading training data, 2) transforming data, 3) training model, and 4) loading validation data. Training a particular model took approximately 11 hours and we were able to leverage HPC techniques in order to simultaneously test a number of competing parameters such as different learning rates and batch sizes for the neural network.

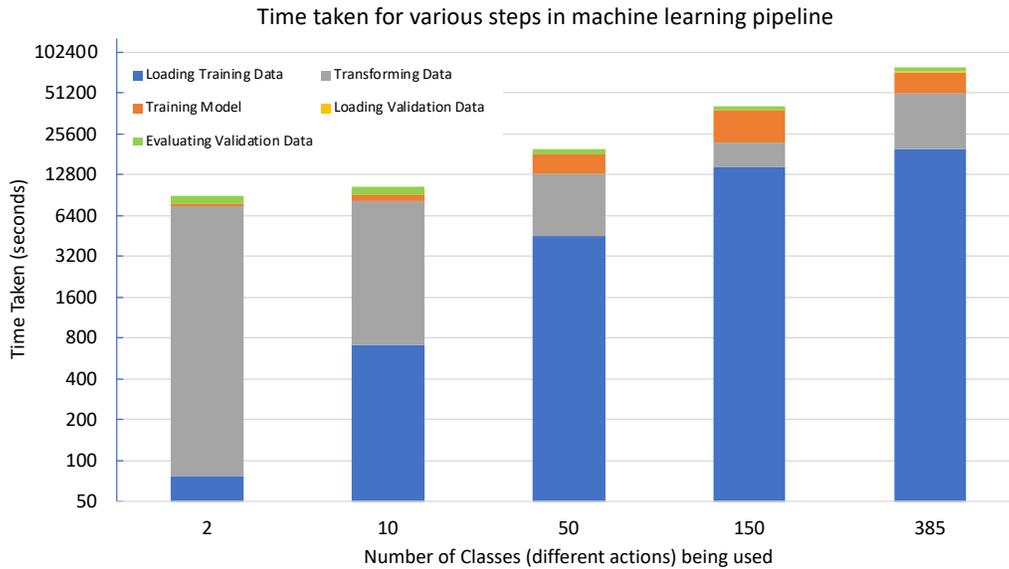

*Figure 4.5. Time taken for different steps on exemplary problem's machine-learning pipeline (note the log scale on the y-axis).*

# 5 ROBUST ARTIFICIAL INTELLIGENCE

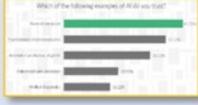

Figure 5.1. Importance of robust AI.

A growing research area critical to the widespread deployment of AI solutions to DoD and IC problems is in the domain of robust AI. We use the term robust AI as a general term that includes explainability, verification/validation, metrics, security (cyber and physical), policy, ethics, safety, and training. Sometimes, this component is also referred to as trusted AI or adversarial AI (not to be confused with GANs—a machine-learning technique). In this section, we describe each of the robust AI features of Figure 5.1. As this field continues to evolve at a rapid pace, we focus on describing the challenges and highlight a few related research results as applicable.

## 5.1 EXPLAINABLE AI

A key aspect of gaining trust in an AI system is in the ability of the system to explain the reasoning behind a particular decision. In particular for DoD/IC applications where AI systems may be supporting complex decisions or decisions with very high impact, it is imperative that users understand how a particular decision was reached by the system. This explanation can help users not only understand the reasoning behind an output, but also may help with the resiliency of algorithms. It is more difficult for an adversary to manipulate a model that needs to explain what it is doing.

Traditional AI systems that rely on expert or knowledge-based systems had the advantage of often being inherently explainable. Since the rules were created by experts and used tools such as decision trees, the AI system could simply output the states that were activated in reaching a decision, and the human user, ostensibly trained to use the system, could simply look at what was activated.

With more recent machine-learning algorithms, however, this is not necessarily the case. In fact, the ability of a machine-learning algorithm to combine multiple features in often non-linear ways is part of the value associated with them! Standard approaches of simply outputting the accuracy or confidence may work for domains in which the cost of an error is low but is unlikely to be sufficient for decision makers in critical applications [111]. Although there are certain machine-learning algorithms such as Bayesian networks that are more amenable to explainability, this property is not necessarily present in most machine-learning techniques. For example, while there is ongoing research [87] looking at the explainability of neural networks, the explainability of most off-the-shelf models is poor. To underscore the importance of this field, the Defense Advanced Research Projects Agency (DARPA) recently began a program associated with explainable AI called XAI [112].

Closely related to explainability is the concept of interpretability. One should be careful with the terms "explainable" and "interpretable" (along with "comprehensible" and "understandable") that are overloaded and often conflated in the AI literature. "Explainable" AI provides an explanation for the AI's recommendation in terms a human can understand, even though the explanation might not fully describe how the AI arrived at its recommendation. The model or process that an AI uses to make its recommendation is said to be "interpretable" if a human can understand it. As examples, consider a neural network and a decision tree. The neural network is typically regarded as an opaque black box whose processing cannot be understood, so it is neither explainable nor interpretable. In contrast, the decision tree follows an explicit sequence of logical steps to make its recommendations, so it is interpretable and hence explainable. One approach to making a neural network explainable is to train another decision tree on examples of the neural network's inputs and outputs and use the new decision tree to approximately describe what the neural network is doing.

For most applications of interest, one will need to develop algorithms that are explainable as well as interpretable. In many cases, providing the explanation of how an algorithm reached a particular decision may only be as good as how well that explanation can be interpreted by the end user. An overly complicated explanation may exceed the capacity of a human end user or domain expert to understand.

## 5.2 METRICS

In discussing metrics, we differentiate between component-level metrics and system-level metrics. Much of the presentation of AI or machine-learning results is done via component-level metrics that provide results on how well a particular component of the AI architecture performed. For example, one may present the accuracy or precision of an algorithm, but this does not indicate how well data conditioning was performed or how much of an impact this made to the overall mission.

Measuring the output of a machine-learning algorithm depends heavily on the task at hand. Within the realm of supervised learning classification, some common measurements include the true positive rate (the number of correct "positive" classifications), the true negative rate (the number of correct "negative" classifications), the false positive rate (the number of incorrect "positive" classifications), and the false negative rate (the number of incorrect "negative" classifications). These measures are often combined to report metrics such as accuracy (the ratio of correct predictions to the total number of samples), precision (the ratio of true positives to true positives and false positives), and recall (the ratio of true positives to true positives and false negatives). Higher level metrics such as the F-score can be used to represent the ratio of precision to recall for example.

The quality of regression can be measured by metrics such as residuals, which measure the algorithms output and compare them with the actual outputs. These residuals can be used to compute metrics such as mean absolute error (the average of absolute values of residuals) or mean square error (the average of the squared values of residuals). In the literature, it is also common to see metrics such as mean absolute percentage error or $R_2$ error.

In the case of unsupervised learning tasks such as clustering, internal structure may be measured by parameters such as the ratio of intra-cluster distances vs. inter-cluster distances, distances between cluster centroids, or mutual information. Other measurements may leverage external information such as ground-truth of cluster labels or known cluster structure [113].

Although there are good component-by-component metrics for the AI canonical architecture, there is a major gap in end-to-end metrics. Thus, one may be able to measure the effectiveness of data conditioning or algorithms or computing but understanding how all of the components work together is a major gap in the presentation of metrics. For example, it is uncertain how one would present the overall impact to a mission by using an AI system. Of course, such metrics would likely need to tie in closely to the mission at hand.

## 5.3 SECURITY

Security researchers look at the confidentiality, integrity, and availability of systems when evaluating threats and defenses [114]. In the same vein, AI researchers will need to evaluate the functioning of their AI system under adversarial conditions. At the core of using AI for important applications is the trust behind the system. Beyond obvious issues such as accuracy and precision of a particular AI pipeline, one must also look at issues that may arise when these systems are used in adversarial settings.

AI applications are prone to numerous attacks that can change the output often in unpredictable ways. For example, an adversary may physically manipulate an image or video that leads to incorrect classification or reduction in confidence in an algorithm. Adversaries may also be able to introduce bias into training data or manipulate sensors collecting data through cyberattacks.

The threat surface of the AI pipeline can be vast and Figure 5.2 describes some of the dimensions that may be used by security researchers in understanding where an attack on their AI system may occur. It should be noted that these dimensions are meant simply to be guidelines and not seen as rigid definitions of the dimensions associated with AI security. Further, the dimensions presented are not perfectly orthogonal to each other, and there are relationships that exist across dimensions.

The first dimension of Figure 5.2—Access—corresponds to what information or knowledge about the AI system an adversary may have access to. This dimension is meant to indicate that an adversary with only limited knowledge and access to an AI system they wish to compromise will be limited in the types of adversarial attacks they can perform. For example, given knowledge of the machine-learning model or architecture used, an adversary may deliberately poison training data such that, under certain circumstances, the model outputs an incorrect classification. There are sophisticated and relatively simple ways of doing this. A simple example of such a data poisoning attack would be to physically manipulate datasets. For example, the authors in [115] were able to fool a machine-learning algorithm into misclassifying a stop sign as a speed limit sign by simply placing a sticker on a stop sign. Although techniques such as those described in [116] and [117] may help protect the confidentiality and/or integrity of training and inference data, sophisticated adversaries can also leverage detailed information such as model architecture, weights, and training tools that may be readily available based on knowledge of public-domain models.

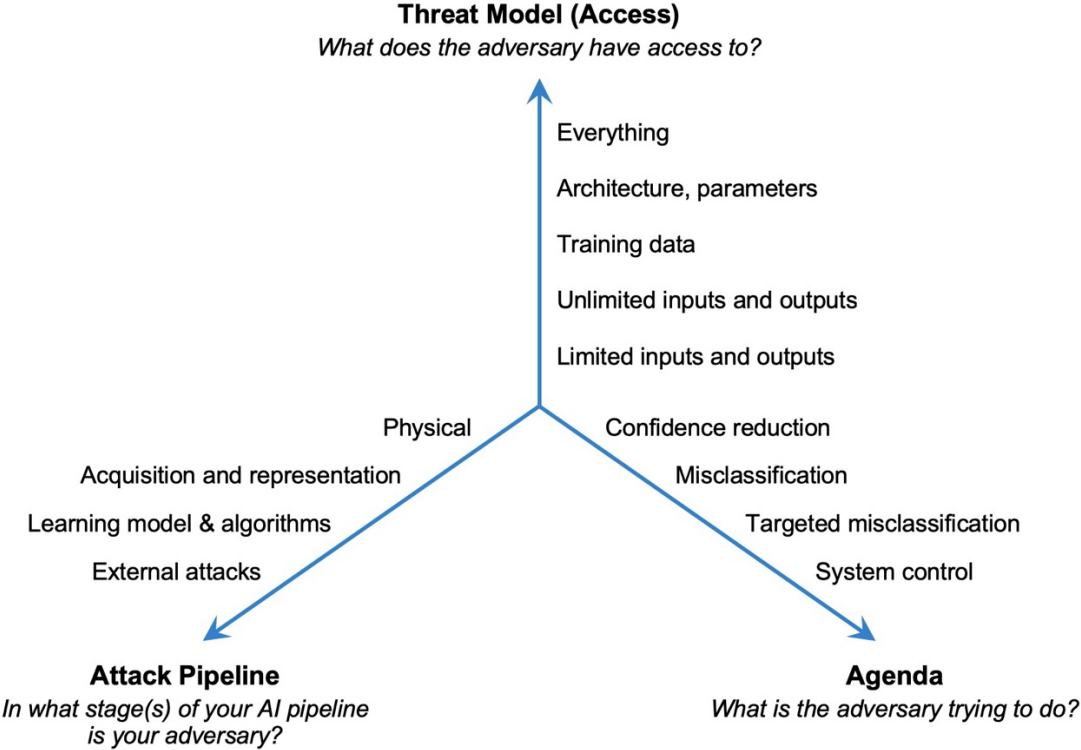

*Figure 5.2. Dimensions of adversarial AI (three A's): access, agenda, attack pipeline.*

In our framework, the second dimension—Agenda—corresponds to what an adversary is trying to achieve with their attack. In a simple case where an adversary is simply trying to reduce the confidence of an algorithm, they may only need access to samples or training data. For more sophisticated attacks that aim for system control, an adversary may need access to many other components or have in-depth knowledge of how a system was designed.

The final dimension in our framework—Attack Pipeline—is meant to indicate that there are different places in the AI pipeline where an attack may occur. Clearly, the types of attacks that can occur on the acquisition and representation of data are different than attacks on the learning model and algorithms. For example, an adversary may try to manipulate a sensor collecting data or attempt to manipulate the weights used in making a determination of a threat.

Again, the presentation of these dimensions is not meant to imply that each of these dimensions are mutually orthogonal or unrelated. Rather, these dimensions are meant to indicate that there are different ways in which an adversary may attempt to infiltrate a system and that there are different capabilities needed to protect against these attacks. There are a number of well-studied and publicized examples of machine-learning algorithms being fooled by seemingly simple attacks [118–120]. Developing algorithms that are robust to such attacks is a very active area of research and there are numerous examples [121–124] of researchers developing counter-measures that can be applied depending on the threat. Beyond attacks specific to the AI pipeline, we should also note that AI system are also prone to a variety of cyber and physical vulnerabilities such as supply-chain risks; data exfiltration attacks; and attacks on the confidentiality, integrity, and/or availability of various components in the AI pipeline. There are a number of active research projects such as [116, 125–127] that aim to provide mitigations for certain classes of attacks.

Similar to many other areas of security research, it is likely that in the near future this field will continue to see a game of "cat-and-mouse" in which security researchers develop new examples of adversarial attacks and develop counter-measures to provide robust AI performance in the face of these attacks. For DoD and IC applications, it is imperative that developers work with security professionals to understand the types of adversarial attacks they may be prone to and develop counter-measures or techniques to minimize the effects of these attacks.

## 5.4  OTHER ROBUST AI FEATURES

Other topics that may need to be studied or developed before widespread deployment of AI solutions to DoD/IC missions include validation and verification and policy, ethics, safety and training. Validation and verification techniques can be used to measure the compliance of various system features with specifications, rules, and conditions under which the system is intended to operate. Although there is limited work in validating and verifying expert systems from a software perspective [128], there is very little current research on applying such techniques to AI systems that leverage more complex machine-learning algorithms such as neural networks.

Finally, widespread deployment of AI systems within the DoD and IC will largely be advanced or impeded by AI rules and regulations. Although there are a number of technical challenges associated with developing such rules and regulations, there will also need to be a consistent effort across multiple agencies to develop best practices and share results.

## 5.5  ROBUST AI WITHIN THE CONTEXT OF EXEMPLARY APPLICATION

Given the research nature of our exemplary application, we did not focus significant effort on security and adversaries. However, we did use simple techniques such as inspecting inputs, outputs, and selective pieces of the neural network to understand where the system was failing and for debugging errors. In measuring results, we use the top-one and top-five accuracies. This metric is defined as follows: An algorithm will label each of the videos with one of $k$ labels. The top-$k$ accuracy says that a video was correctly identified if one of its top $k$ labels are the correct label. For example, a video may be classified (in decreasing probability) as: *(barking, yelling, running, …)*. If the correct label (as judged by a human observer) is "yelling", this would contribute a correct classification toward the top-five but would contribute a miss towards the top-one accuracy. As of June 2018, leading models for the Moments in Time Dataset had top-one accuracies of approximately 0.3 and top-five accuracies of approximately 0.6 [3, 5].

# 6    HUMAN–MACHINE TEAMING

The final piece of the canonical AI architecture of Figure 1.1 is what we refer to as human–machine teaming. This piece is critical in connecting the AI system to the end user and mission. Human–machine teaming tasks are defined as tasks in which the human and machine system are interdependent in some fashion. Human–machine collaboration is a broader defined term that encompasses interdependent tasks, but also tasks that are not interdependent such as sequential or loosely coordinated tasks.

First of all, it is important to understand which tasks are mapped well to humans and which tasks are mapped well to machines. As shown in Figure 6.1, there is a spectrum that relate to how closely humans and machines work together. Borrowing the terminology from [129], we refer to broad collaboration relationships as human-in-the-loop, human-on-the-loop, and human-out-of-the-loop. Human-in-the-loop collaboration is when a human is closely in the loop of the AI system. For example, a human and machine working together to jointly solve a common goal. In this relationship, the human and machine are equal contributors (or in certain cases, the human is a greater participant) to the overall system. The second type of relationship is referred to as human-on-the-loop. In this form of interaction, a human is largely participating in the system in a supervisory capacity. Human-on-the-loop systems will largely leverage automated techniques and may triage more important information to a human observer or the human may provide oversight on the functioning of the system. The final type of relationship is referred to as human-out-of-the-loop. In this relationship, the human does not participate in the AI system's operation under normal conditions.

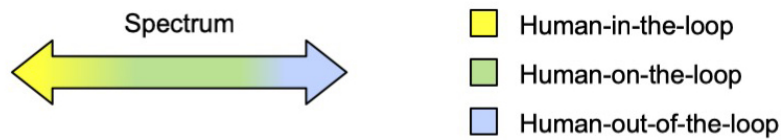

*Figure 6.1. Spectrum of humans and machines interacting.*

Clearly, different applications will have different requirements in terms of how closely humans and machine work together. In order to determine the appropriate level of human and machine interaction, Figure 6.2 describes a high-level framework that may assist in such a mapping. On the horizontal axis, we look at the consequence of actions—how important is it that the system provide the correct response. On the vertical axis, we look at the confidence in the machine making the decision. Clearly, for very consequential decisions that may impact lives in a significant way, such decisions are clearly mapped well to humans. On the other hand, for low consequence decisions in which we have high confidence in the machine, such decisions may be well mapped to machines. Within the context of the DoD and IC applications, in the near future, it is likely that AI systems will largely be used to augment human decision making. This is largely due to the high consequence of decisions. Certain tasks such as anomaly detection or highlighting important data may be performed by a system but high-consequence decisions such as whether to deploy troops or resources will likely still be performed by humans. Historically, systems have been designed with a static human/machine task allocation, but in current and future systems, task authority can dynamically change from human to machine depending on the context or the capabilities of the human or machine.

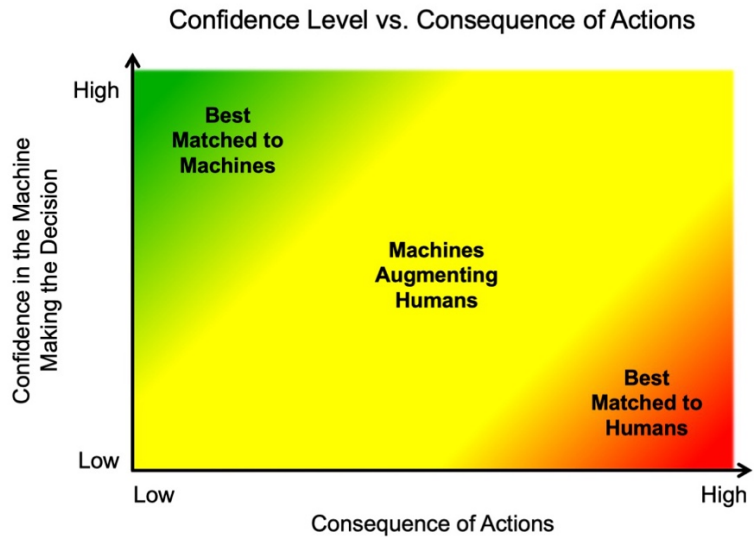

*Figure 6.2. Determining which tasks map well to humans and machines.*

For those systems in which the humans and machines must interact to be successful, it is important to provide the elements to enable an effective human–machine collaboration. In Figure 6.3, some of these elements are outlined. At the top are environmental elements that provide part of the context in which a collaboration occurs. There are static elements such as physics of the environment, which do not change dynamically. There are semi-static elements such as physical infrastructure including buildings, rivers, and forests. There are social constructs that provide limitations on and rules about how people interact with one another (and these could vary depending on what part of the globe one is in). Mission provides the other part of the context. Goals of the mission establish why the human–machine collaboration is taking place. Tasks outline how these goals can be accomplished. Procedures are organizationally determined structure on how the tasks are performed.

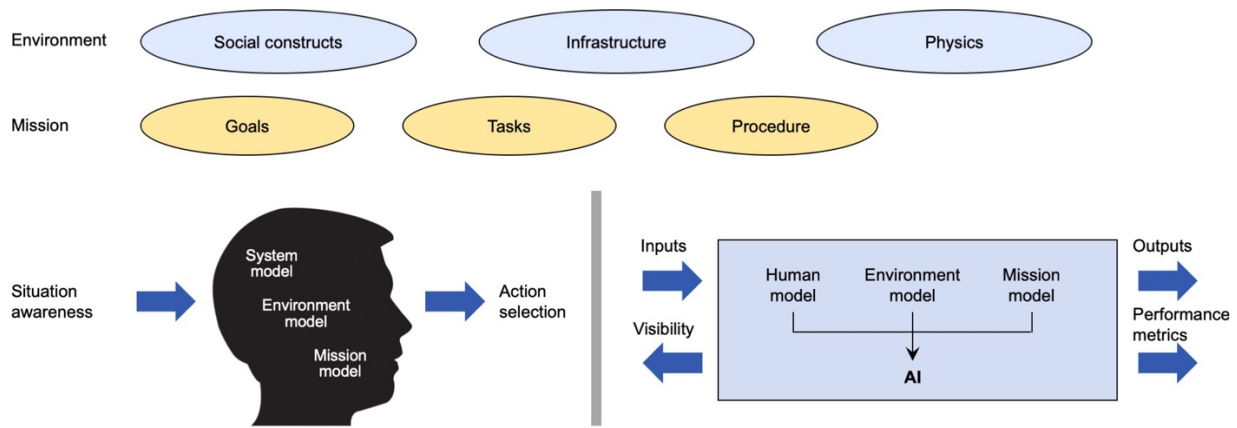

*Figure 6.3. Elements of effective human–machine collaboration.*

The humans then perceive information from the context and from the machine system. Using this perceived information, the human projects expected behavior for the system, the environment, and the mission into the future. With these predictions, course of action decisions can be made.

The system also receives input from the human and the environment. Ideally, the system would contain models of the human, environment, and mission in order to extrapolate expected behavior into the future. These predictions could then be used to make system behavior decisions. In reality, much of the human–machine collaboration issues stem from the fact that the human, environment, and mission models within the system are faulty or non-existent. Systems can also have more or less information visible to the human as it continues its tasks. As systems lean towards being adaptive AI, to ensure that the system model within the human is functionally accurate, more visibility into the system's processes should be provided. Likewise, the more information the system has about the human's evolving tasks and models, the better the system's decisions will be.

Johnson's research [130] has identified three aspects of the interaction between humans and machines that establish effective human–machine teaming: observability, predictability, and directability. Observability is akin to the "visibility" description above—can the human and machine see what the other is doing/planning to do? Predictability is the capability resulting from the models within the human/machine. If you have a model of the human or system or environment, you can then predict the future behavior. Directability is the means of being able to exert control on the human/system teammate. If one of the human's courses of action were to tell the system what to do (or vice versa), then the interaction would have the quality of directability. Often directability lies on a spectrum from low directability (infrequent or gross control) to high directability (continuous and fine control). Outstanding research questions remain on how much observability, predictability, and directability is required for a sufficient teaming interaction for different tasks. One interesting finding from Johnson's research is that as the system becomes more autonomous, MORE attention to human–machine collaboration, not less, is required to maintain complete system performance.

There are a number of research areas such as data visualization and algorithm interpretability that may also help move these boundaries. Although there has been research on how humans and machines interact in domains such as robotics [131, 132], research on the interaction between humans and machines for more general AI applications is relatively limited. In the near future it seems that AI systems will need to be developed in close collaboration with end users in order to design how systems interact with the end users.

There are a number of techniques in the field of human–computer interaction that may be used as a starting point for researchers and developers. For example, principles of user design, usability and interface testing. New visualization research on augmented reality, 3D printing, immersive gaming environments, and brain–computer interfaces may also play a large part in developing novel human–machine teaming interfaces.

For the DoD and IC, the possibilities of humans and machine working together are tremendous. Although certain tasks will likely continue to be human-only, there are a number of possibilities for applications human-in-the-loop and human-on-the-loop relationships. For example, in [129], the authors highlight AI roles such as training, interacting, and amplifying as prime candidates for either humans complementing machine intelligence or AI providing humans with superpowers (as described by P.R. Daugherty in Figure P2-1 on page 107, in *Human+ Machine: Reimagining Work in the Age of AI* [129]). Certain mundane, cumbersome, or trivial tasks may also be candidates for human-out-of-the-loop solutions.

# 7 ACKNOWLEDGEMENTS

The authors wish to thank the following individuals for their contributions, thoughts, and comments toward the development of this section: Charlie Dagli, Arjun Majumdar, Lauren Milechin, Julia Mullen, Jonathan Herrera, Paul Monticciolo, Nicolas Malyska, and William Streilein, Additional thanks to Laura Glazer and Brad Dillman for their editorial help.

# 8 REFERENCES

[1] T. N. Theis and H.-S. P. Wong, "The end of moore's law: A new beginning for information technology," *Computing in Science & Engineering,* vol. 19, no. 2, pp. 41-50, 2017.

[2] M. Monfort *et al.*, "Moments in Time Dataset: one million videos for event understanding," *arXiv preprint arXiv:1801.03150,* 2018.

[3] C. Li *et al.*, "Team DEEP-HRI Moments in Time Challenge 2018 Technical Report."

[4] Y. Li *et al.*, "Submission to Moments in Time Challenge 2018."

[5] Z. Xiaoteng *et al.*, "Qiniu Submission to ActivityNet Challenge 2018."

[6] G. Press, "Cleaning big data: Most time-consuming, least enjoyable data science task, survey says," *Forbes, March,* vol. 23, 2016.

[7] A. Krizhevsky, I. Sutskever, and G. Hinton, "ImageNet classification with deep convolutional neural networks. Advances in Neural Information Processing Systems 25 (NIPS 2012)," ed, 2017.

[8] Y. LeCun, C. Cortes, and C. Burges, "MNIST handwritten digit database," *AT&T Labs [Online]. Available: http://yann. lecun. com/exdb/mnist,* vol. 2, 2010.

[9] A. Krizhevsky, V. Nair, and G. Hinton, "The CIFAR-10 dataset," *online: http://www. cs. toronto. edu/kriz/cifar. html,* 2014.

[10] V. Kurin, S. Nowozin, K. Hofmann, L. Beyer, and B. Leibe, "The Atari Grand Challenge Dataset," *arXiv preprint arXiv:1705.10998,* 2017.

[11] P. Borgnat, G. Dewaele, K. Fukuda, P. Abry, and K. Cho, "Seven years and one day: Sketching the evolution of internet traffic," in *INFOCOM 2009, IEEE*, 2009, pp. 711-719: IEEE.

[12] P. J. Braam, "The Lustre storage architecture," 2004.

[13] K. Loney, *Oracle Database 11g The Complete Reference*. McGraw-Hill, Inc., 2008.

[14] M. Stonebraker and L. A. Rowe, *The design of Postgres* (no. 2). ACM, 1986.

[15] S. Gilbert and N. Lynch, "Brewer's conjecture and the feasibility of consistent, available, partition-tolerant web services," *Acm Sigact News,* vol. 33, no. 2, pp. 51-59, 2002.

[16] V. Gadepally *et al.*, "The bigdawg polystore system and architecture," in *High Performance Extreme Computing Conference (HPEC), 2016 IEEE*, 2016, pp. 1-6: IEEE.

[17] F. Chang *et al.*, "Bigtable: A distributed storage system for structured data," *ACM Transactions on Computer Systems (TOCS),* vol. 26, no. 2, p. 4, 2008.

[18] A. Lakshman and P. Malik, "Cassandra: a decentralized structured storage system," *ACM SIGOPS Operating Systems Review,* vol. 44, no. 2, pp. 35-40, 2010.

[19] J. Han, E. Haihong, G. Le, and J. Du, "Survey on NoSQL database," in *Pervasive computing and applications (ICPCA), 2011 6th international conference on*, 2011, pp. 363-366: IEEE.

[20] J. Kepner *et al.*, "Achieving 100,000,000 database inserts per second using Accumulo and D4M," in *High Performance Extreme Computing Conference (HPEC), 2014 IEEE*, 2014, pp. 1-6: IEEE.

[21] M. Stonebraker, "Newsql: An alternative to nosql and old sql for new oltp apps," *Communications of the ACM. Retrieved,* pp. 07-06, 2012.

[22] J. Kepner et al., "Associative array model of SQL, NoSQL, and NewSQL databases," in *2016 IEEE High Performance Extreme Computing Conference (HPEC)*, 2016, pp. 1-9: IEEE.

[23] J. Kepner and H. Jananthan, *Mathematics of big data: Spreadsheets, databases, matrices, and graphs*. MIT Press, 2018.

[24] R. Tan, R. Chirkova, V. Gadepally, and T. G. Mattson, "Enabling query processing across heterogeneous data models: A survey," in *Big Data (Big Data), 2017 IEEE International Conference on*, 2017, pp. 3211-3220: IEEE.

[25] A. Elmore et al., "A demonstration of the bigdawg polystore system," *Proceedings of the VLDB Endowment,* vol. 8, no. 12, pp. 1908-1911, 2015.

[26] T. Mattson, V. Gadepally, Z. She, A. Dziedzic, and J. Parkhurst, "Demonstrating the BigDAWG Polystore System for Ocean Metagenomics Analysis," in *CIDR*, 2017.

[27] P. Chen, V. Gadepally, and M. Stonebraker, "The bigdawg monitoring framework," in *2016 IEEE High Performance Extreme Computing Conference (HPEC)*, 2016, pp. 1-6: IEEE.

[28] Z. She, S. Ravishankar, and J. Duggan, "Bigdawg polystore query optimization through semantic equivalences," in *2016 IEEE High Performance Extreme Computing Conference (HPEC)*, 2016, pp. 1-6: IEEE.

[29] A. M. Gupta, V. Gadepally, and M. Stonebraker, "Cross-engine query execution in federated database systems," in *2016 IEEE High Performance Extreme Computing Conference (HPEC)*, 2016, pp. 1-6: IEEE.

[30] A. Dziedzic, A. J. Elmore, and M. Stonebraker, "Data transformation and migration in polystores," in *2016 IEEE High Performance Extreme Computing Conference (HPEC)*, 2016, pp. 1-6: IEEE.

[31] D. Deng et al., "The Data Civilizer System," in *CIDR*, 2017.

[32] Z. Abedjan, J. Morcos, I. F. Ilyas, M. Ouzzani, P. Papotti, and M. Stonebraker, "Dataxformer: A robust transformation discovery system," in *2016 IEEE 32nd International Conference on Data Engineering (ICDE)*, 2016, pp. 1134-1145: IEEE.

[33] M. Stonebraker et al., "Data Curation at Scale: The Data Tamer System," in *CIDR*, 2013.

[34] S. Kandel, A. Paepcke, J. Hellerstein, and J. Heer, "Wrangler: Interactive visual specification of data transformation scripts," in *Proceedings of the SIGCHI Conference on Human Factors in Computing Systems*, 2011, pp. 3363-3372: ACM.

[35] V. Hodge and J. Austin, "A survey of outlier detection methodologies," *Artificial intelligence review,* vol. 22, no. 2, pp. 85-126, 2004.

[36] V. Gadepally and J. Kepner, "Big data dimensional analysis," in *2014 IEEE High Performance Extreme Computing Conference (HPEC)*, 2014, pp. 1-6: IEEE.

[37] V. Gadepally and J. Kepner, "Using a power law distribution to describe big data," in *2015 IEEE High Performance Extreme Computing Conference (HPEC)*, 2015, pp. 1-5: IEEE.

[38] A. Krizhevsky, I. Sutskever, and G. E. Hinton, "Imagenet classification with deep convolutional neural networks," in *Advances in neural information processing systems*, 2012, pp. 1097-1105.

[39] B. Zhou, A. Lapedriza, J. Xiao, A. Torralba, and A. Oliva, "Learning deep features for scene recognition using places database," in *Advances in neural information processing systems*, 2014, pp. 487-495.

[40] T.-Y. Lin *et al.*, "Microsoft coco: Common objects in context," in *European conference on computer vision*, 2014, pp. 740-755: Springer.

[41] G. Paolacci, J. Chandler, and P. G. Ipeirotis, "Running experiments on amazon mechanical turk," 2010.

[42] A. Sorokin and D. Forsyth, "Utility data annotation with amazon mechanical turk," in *Computer Vision and Pattern Recognition Workshops, 2008. CVPRW'08. IEEE Computer Society Conference on*, 2008, pp. 1-8: IEEE.

[43] P. G. Ipeirotis, F. Provost, and J. Wang, "Quality management on amazon mechanical turk," in *Proceedings of the ACM SIGKDD workshop on human computation*, 2010, pp. 64-67: ACM.

[44] A. Dobkin. (2017). *DOD Maven AI project develops first algorithms, starts testing*. Available: https://defensesystems.com/articles/2017/11/03/maven-dod.aspx

[45] T. N. Mundhenk, G. Konjevod, W. A. Sakla, and K. Boakye, "A large contextual dataset for classification, detection and counting of cars with deep learning," in *European Conference on Computer Vision*, 2016, pp. 785-800: Springer.

[46] C. Vondrick, A. Shrivastava, A. Fathi, S. Guadarrama, and K. Murphy, "Tracking emerges by colorizing videos," *arXiv preprint arXiv:1806.09594,* 2018.

[47] J. Lafferty, A. McCallum, and F. C. Pereira, "Conditional random fields: Probabilistic models for segmenting and labeling sequence data," 2001.

[48] I. Goodfellow, Y. Bengio, A. Courville, and Y. Bengio, *Deep learning*. MIT press Cambridge, 2016.

[49] B. G. Buchanan and E. A. Feigenbaum, "DENDRAL and Meta-DENDRAL: Their applications dimension," in *Readings in artificial intelligence*: Elsevier, 1981, pp. 313-322.

[50] F. Hayes-Roth, D. A. Waterman, and D. B. Lenat, "Building expert system," 1983.

[51] D. Waterman, "A guide to expert systems," 1986.

[52] A. J. Gonzalez and D. D. Dankel, *The engineering of knowledge-based systems*. Prentice-Hall Englewood Cliffs, NJ, 1993.

[53] V. Gadepally, A. Kurt, A. Krishnamurthy, and Ü. Özgüner, "Driver/vehicle state estimation and detection," in *Intelligent Transportation Systems (ITSC), 2011 14th International IEEE Conference on*, 2011, pp. 582-587: IEEE.

[54] V. Gadepally, A. Krishnamurthy, and U. Ozguner, "A framework for estimating driver decisions near intersections," *IEEE Transactions on Intelligent Transportation Systems,* vol. 15, no. 2, pp. 637-646, 2014.

[55] P. Domingos, *The master algorithm: How the quest for the ultimate learning machine will remake our world*. Basic Books, 2015.

[56] L. E. Peterson, "K-nearest neighbor," *Scholarpedia,* vol. 4, no. 2, p. 1883, 2009.

[57] A. Paszke *et al.*, "Automatic differentiation in pytorch," 2017.

[58] M. Abadi *et al.*, "Tensorflow: a system for large-scale machine learning," in *OSDI*, 2016, vol. 16, pp. 265-283.

[59] Y. Jia *et al.*, "Caffe: Convolutional architecture for fast feature embedding," in *Proceedings of the 22nd ACM international conference on Multimedia*, 2014, pp. 675-678: ACM.

[60] R. Socher, C. C. Lin, C. Manning, and A. Y. Ng, "Parsing natural scenes and natural language with recursive neural networks," in *Proceedings of the 28th international conference on machine learning (ICML-11)*, 2011, pp. 129-136.

[61] G. E. Hinton, "Deep belief networks," *Scholarpedia,* vol. 4, no. 5, p. 5947, 2009.

[62] B. Karlik and A. V. Olgac, "Performance analysis of various activation functions in generalized MLP architectures of neural networks," *International Journal of Artificial Intelligence and Expert Systems,* vol. 1, no. 4, pp. 111-122, 2011.

[63] I. Goodfellow et al., "Generative adversarial nets," in *Advances in neural information processing systems*, 2014, pp. 2672-2680.

[64] M. Chan, D. Scarafoni, R. Duarte, J. Thornton, and L. Skelly, "Learning Network Architectures of Deep CNNs under Resource Constraints," in *Proceedings of the IEEE Conference on Computer Vision and Pattern Recognition Workshops*, 2018, pp. 1703-1710.

[65] C. Szegedy, S. Ioffe, V. Vanhoucke, and A. A. Alemi, "Inception-v4, inception-resnet and the impact of residual connections on learning," in *AAAI*, 2017, vol. 4, p. 12.

[66] A. Parashar et al., "Scnn: An accelerator for compressed-sparse convolutional neural networks," in *2017 ACM/IEEE 44th Annual International Symposium on Computer Architecture (ISCA)*, 2017, pp. 27-40: IEEE.

[67] V. Gadepally, J. Kepner, and A. Reuther, "Storage and database management for big data," *Big Data: Storage, Sharing and Security. CRC Press. Retrieved on July,* vol. 30, p. 2017, 2016.

[68] X. Zhu, "Semi-supervised learning literature survey," *Computer Science, University of Wisconsin-Madison,* vol. 2, no. 3, p. 4, 2006.

[69] X. Zhu, "Semi-supervised learning," in *Encyclopedia of machine learning*: Springer, 2011, pp. 892-897.

[70] R. S. Sutton and A. G. Barto, *Reinforcement learning: An introduction*. MIT press, 2018.

[71] J. Kober and J. Peters, "Reinforcement learning in robotics: A survey," in *Reinforcement Learning*: Springer, 2012, pp. 579-610.

[72] V. Mnih et al., "Playing atari with deep reinforcement learning," *arXiv preprint arXiv:1312.5602,* 2013.

[73] D. Silver et al., "Mastering the game of Go without human knowledge," *Nature,* vol. 550, no. 7676, p. 354, 2017.

[74] P. Abbeel, A. Coates, M. Quigley, and A. Y. Ng, "An application of reinforcement learning to aerobatic helicopter flight," in *Advances in neural information processing systems*, 2007, pp. 1-8.

[75] N. Jouppi, "Google supercharges machine learning tasks with TPU custom chip," *Google Blog, May,* vol. 18, 2016.

[76] M. L. Minsky, *Computation: finite and infinite machines*. Prentice-Hall, Inc., 1967.

[77] N. P. Jouppi, C. Young, N. Patil, and D. Patterson, "A domain-specific architecture for deep neural networks," *Communications of the ACM,* vol. 61, no. 9, pp. 50-59, 2018.

[78] J. Dean, D. Patterson, and C. Young, "A New Golden Age in Computer Architecture: Empowering the Machine-Learning Revolution," *IEEE Micro,* vol. 38, no. 2, pp. 21-29, 2018.

[79] P. A. Merolla et al., "A million spiking-neuron integrated circuit with a scalable communication network and interface," *Science,* vol. 345, no. 6197, pp. 668-673, 2014.

[80]     R. R. Schaller, "Moore's law: past, present and future," *IEEE spectrum,* vol. 34, no. 6, pp. 52-59, 1997.

[81]     D. J. Frank, R. H. Dennard, E. Nowak, P. M. Solomon, Y. Taur, and H.-S. P. Wong, "Device scaling limits of Si MOSFETs and their application dependencies," *Proceedings of the IEEE,* vol. 89, no. 3, pp. 259-288, 2001.

[82]     OpenAI. (October 27, 2018). *AI and Compute*. Available: https://blog.openai.com/ai-and-compute/

[83]     N. P. Jouppi *et al.*, "In-datacenter performance analysis of a tensor processing unit," in *Computer Architecture (ISCA), 2017 ACM/IEEE 44th Annual International Symposium on*, 2017, pp. 1-12: IEEE.

[84]     N. Hemsoth. (2017). *Startup's AI Chip Beats GPU*. Available: https://www.nextplatform.com/2017/08/23/first-depth-view-wave-computings-dpu-architecture-systems/

[85]     D. Lacey. (2017). *Preliminary IPU Benchmarks*. Available: https://www.graphcore.ai/posts/preliminary-ipu-benchmarks-providing-previously-unseen-performance-for-a-range-of-machine-learning-applications

[86]     R. Merritt. (2018). *Startup's AI Chip Beats GPU*. Available: https://www.eetimes.com/document.asp?doc_id=1333719

[87]     C. Chen, O. Li, A. Barnett, J. Su, and C. Rudin, "This looks like that: deep learning for interpretable image recognition," *arXiv preprint arXiv:1806.10574,* 2018.

[88]     P. Micikevicius *et al.*, "Mixed precision training," *arXiv preprint arXiv:1710.03740,* 2017.

[89]     S. Gupta, A. Agrawal, K. Gopalakrishnan, and P. Narayanan, "Deep learning with limited numerical precision," in *International Conference on Machine Learning*, 2015, pp. 1737-1746.

[90]     T. Chen *et al.*, "Mxnet: A flexible and efficient machine learning library for heterogeneous distributed systems," *arXiv preprint arXiv:1512.01274,* 2015.

[91]     F. Seide and A. Agarwal, "CNTK: Microsoft's open-source deep-learning toolkit," in *Proceedings of the 22nd ACM SIGKDD International Conference on Knowledge Discovery and Data Mining*, 2016, pp. 2135-2135: ACM.

[92]     J. Keuper and F.-J. Preundt, "Distributed training of deep neural networks: theoretical and practical limits of parallel scalability," presented at the Proceedings of the Workshop on Machine Learning in High Performance Computing Environments, Salt Lake City, Utah, 2016.

[93]     T. Domhan, J. T. Springenberg, and F. Hutter, "Speeding Up Automatic Hyperparameter Optimization of Deep Neural Networks by Extrapolation of Learning Curves," in *IJCAI*, 2015, vol. 15, pp. 3460-8.

[94]     P. R. Lorenzo, J. Nalepa, M. Kawulok, L. S. Ramos, and J. R. Pastor, "Particle swarm optimization for hyper-parameter selection in deep neural networks," in *Proceedings of the Genetic and Evolutionary Computation Conference*, 2017, pp. 481-488: ACM.

[95]     P. Goyal *et al.*, "Accurate, Large Minibatch SGD: Training ImageNet in 1 Hour," ed, 2017.

[96]     T. Kurth *et al.*, "Deep learning at 15pf: supervised and semi-supervised classification for scientific data," in *Proceedings of the International Conference for High Performance Computing, Networking, Storage and Analysis*, 2017, p. 7: ACM.

[97] Y. You, A. Buluç, and J. Demmel, "Scaling deep learning on GPU and knights landing clusters," in *Proceedings of the International Conference for High Performance Computing, Networking, Storage and Analysis*, 2017, p. 9: ACM.

[98] J. Hines, "Stepping up to Summit," *Computing in Science & Engineering,* vol. 20, no. 2, pp. 78-82, 2018.

[99] V. Gadepally *et al.*, "D4m: Bringing associative arrays to database engines," in *High Performance Extreme Computing Conference (HPEC), 2015 IEEE*, 2015, pp. 1-6: IEEE.

[100] A. Prout *et al.*, "MIT SuperCloud portal workspace: Enabling HPC web application deployment," *arXiv preprint arXiv:1707.05900,* 2017.

[101] R. Chandra, L. Dagum, D. Kohr, R. Menon, D. Maydan, and J. McDonald, *Parallel programming in OpenMP*. Morgan kaufmann, 2001.

[102] A. B. Yoo, M. A. Jette, and M. Grondona, "Slurm: Simple linux utility for resource management," in *Workshop on Job Scheduling Strategies for Parallel Processing*, 2003, pp. 44-60: Springer.

[103] W. Gentzsch, "Sun grid engine: Towards creating a compute power grid," in *Proceedings First IEEE/ACM International Symposium on Cluster Computing and the Grid*, 2001, pp. 35-36: IEEE.

[104] B. Hindman *et al.*, "Mesos: A platform for fine-grained resource sharing in the data center," in *NSDI*, 2011, vol. 11, no. 2011, pp. 22-22.

[105] V. K. Vavilapalli *et al.*, "Apache hadoop yarn: Yet another resource negotiator," in *Proceedings of the 4th annual Symposium on Cloud Computing*, 2013, p. 5: ACM.

[106] J. Kepner *et al.*, "Lustre, hadoop, accumulo," in *2015 IEEE High Performance Extreme Computing Conference (HPEC)*, 2015, pp. 1-5: IEEE.

[107] F. Chollet, "Keras," ed, 2015.

[108] A. Prout *et al.*, "Enabling on-demand database computing with MIT SuperCloud database management system," *arXiv preprint arXiv:1506.08506,* 2015.

[109] A. Reuther *et al.*, "LLSuperCloud: Sharing HPC systems for diverse rapid prototyping," 2013 IEEE High Performance Extreme Computing Conference (HPEC), Sept. 10-12, 2013, 2013.

[110] A. Reuther *et al.*, "Interactive supercomputing on 40,000 cores for machine learning and data analysis," in *2018 IEEE High Performance extreme Computing Conference (HPEC)*, 2018, pp. 1-6: IEEE.

[111] O. Biran and C. Cotton, "Explanation and justification in machine learning: A survey," in *IJCAI-17 Workshop on Explainable AI (XAI)*, 2017, p. 8.

[112] D. Gunning, "Explainable artificial intelligence (xai)," *Defense Advanced Research Projects Agency (DARPA), nd Web,* 2017.

[113] K. Wang, B. Wang, and L. Peng, "CVAP: validation for cluster analyses," *Data Science Journal,* vol. 8, pp. 88-93, 2009.

[114] B. Fuller *et al.*, "Sok: Cryptographically protected database search," in *Security and Privacy (SP), 2017 IEEE Symposium on*, 2017, pp. 172-191: IEEE.

[115] I. Evtimov *et al.*, "Robust physical-world attacks on machine learning models," *arXiv preprint arXiv:1707.08945,* vol. 2, no. 3, p. 4, 2017.

[116] V. Gadepally *et al.*, "Computing on masked data to improve the security of big data," in *2015 IEEE International Symposium on Technologies for Homeland Security (HST)*, 2015, pp. 1-6: IEEE.

[117] B. Fuller *et al.*, "Sok: Cryptographically protected database search," in *2017 IEEE Symposium on Security and Privacy (SP)*, 2017, pp. 172-191: IEEE.

[118] N. Carlini and D. Wagner, "Adversarial examples are not easily detected: Bypassing ten detection methods," in *Proceedings of the 10th ACM Workshop on Artificial Intelligence and Security*, 2017, pp. 3-14: ACM.

[119] A. Kurakin, I. Goodfellow, and S. Bengio, "Adversarial examples in the physical world," *arXiv preprint arXiv:1607.02533,* 2016.

[120] S.-M. Moosavi-Dezfooli, A. Fawzi, and P. Frossard, "Deepfool: a simple and accurate method to fool deep neural networks," in *Proceedings of the IEEE Conference on Computer Vision and Pattern Recognition*, 2016, pp. 2574-2582.

[121] J. Lu, T. Issaranon, and D. A. Forsyth, "SafetyNet: Detecting and Rejecting Adversarial Examples Robustly," in *ICCV*, 2017, pp. 446-454.

[122] D. Meng and H. Chen, "Magnet: a two-pronged defense against adversarial examples," in *Proceedings of the 2017 ACM SIGSAC Conference on Computer and Communications Security*, 2017, pp. 135-147: ACM.

[123] N. Papernot, P. McDaniel, X. Wu, S. Jha, and A. Swami, "Distillation as a defense to adversarial perturbations against deep neural networks," in *2016 IEEE Symposium on Security and Privacy (SP)*, 2016, pp. 582-597: IEEE.

[124] U. Shaham, Y. Yamada, and S. Negahban, "Understanding adversarial training: Increasing local stability of neural nets through robust optimization," *arXiv preprint arXiv:1511.05432,* 2015.

[125] R. A. Popa, C. Redfield, N. Zeldovich, and H. Balakrishnan, "CryptDB: protecting confidentiality with encrypted query processing," in *Proceedings of the Twenty-Third ACM Symposium on Operating Systems Principles*, 2011, pp. 85-100: ACM.

[126] C. Juvekar, V. Vaikuntanathan, and A. Chandrakasan, "{GAZELLE}: A Low Latency Framework for Secure Neural Network Inference," in *27th {USENIX} Security Symposium ({USENIX} Security 18)*, 2018, pp. 1651-1669.

[127] R. Gilad-Bachrach, N. Dowlin, K. Laine, K. Lauter, M. Naehrig, and J. Wernsing, "Cryptonets: Applying neural networks to encrypted data with high throughput and accuracy," in *International Conference on Machine Learning*, 2016, pp. 201-210.

[128] A. Vermesan and F. Coenen, *Validation and Verification of Knowledge Based Systems: Theory, Tools and Practice*. Springer Science & Business Media, 2013.

[129] P. R. Daugherty and H. J. Wilson, *Human+ Machine: Reimagining Work in the Age of AI*. Harvard Business Press, 2018.

[130] M. Johnson, J. M. Bradshaw, P. J. Feltovich, C. M. Jonker, M. B. Van Riemsdijk, and M. Sierhuis, "Coactive design: Designing support for interdependence in joint activity," *Journal of Human-Robot Interaction,* vol. 3, no. 1, pp. 43-69, 2014.

[131] M. A. Goodrich and A. C. Schultz, "Human–robot interaction: a survey," *Foundations and Trends® in Human–Computer Interaction,* vol. 1, no. 3, pp. 203-275, 2008.

[132] R. Parasuraman, M. Barnes, K. Cosenzo, and S. Mulgund, "Adaptive automation for human-robot teaming in future command and control systems," Army research lab aberdeen proving ground md human research and engineering directorate2007.